\documentclass[11pt]{article}
\PassOptionsToPackage{hyperfootnotes=false}{hyperref}
\usepackage{acl}
\usepackage{times}
\usepackage{latexsym}
\usepackage{array,amsmath}
\usepackage{booktabs,tabularx,amsmath}
\newcolumntype{Y}{>{\raggedright\arraybackslash}X}
\usepackage{amssymb}
\usepackage{makecell}
\usepackage{xcolor}
\usepackage{setspace}
\usepackage{amsmath}
\usepackage{xspace}
\usepackage{multirow}

\usepackage{tikz}
\usetikzlibrary{positioning, arrows.meta, shapes.geometric, shapes.misc}
\usepackage{placeins}

\newcommand{\our}{\text{Sci-ZSEL}\xspace}
\usepackage[T1]{fontenc}
\usepackage[utf8]{inputenc}
\usepackage{microtype}

\usepackage{graphicx}

\title{Bridging Lexical Divergence:  LLM-Assisted, Cost-Efficient, Zero-shot Scientific Entity Linking }

\author{
Md Rasel Khondokar\textsuperscript{*}, Qiao Qiao\textsuperscript{*}, Farjana Sultana Samia, Nhat Le, Yuepei Li, Qi Li \\
Department of Computer Science, Iowa State University, Ames, Iowa, USA \\
\texttt{\{rasel, qqiao1, fssamia, lan0908, liyp0095, qli\}@iastate.edu}
}

\begin{document}
\maketitle

\begingroup
\renewcommand{\thefootnote}{\fnsymbol{footnote}}
\footnotetext[1]{Equal contribution.}
\endgroup

\begin{abstract} 
Scientific domain entity linking (EL) differs from general domain EL because mentions and entity names often lack lexical overlap. Another challenge is that specialized terminology is used in the scientific domain, which is rarely encountered in models pretrained on general domains. Therefore, models trained on general domains transfer poorly to scientific domains. To address this, in-domain fine-tuning is the natural remedy. However, many scientific domains lack expert-annotated data, motivating the need for a zero-human-annotation approach. Existing zero-shot methods heavily rely on LLMs to generate aliases across entire mention corpora, which incurs substantial computational cost, and those methods provide no mechanism to filter out noise from LLMs.
To address these challenges, we propose \our \footnote{Code and data: \url{https://github.com/rasel-isu/Sci-ZSEL}}, a framework that selectively generates entity aliases with an LLM to control computational cost, and applies an ontology-aware filter to remove aliases that semantically drift toward ontology neighbors. Then, filtered aliases are used to construct pseudo-labeled mention-entity pairs for fine-tuning. To enable evaluation of EL under low lexical overlap, we also release a new animal science EL benchmark linked to three livestock trait ontologies, where mentions and entities exhibit substantially lower lexical overlap than in existing benchmarks. Across five benchmarks, \our outperforms the non-fine-tuned baseline, is most useful on non-overlapping mentions, and combining it with curated synonyms gives the best performance in most settings.
\end{abstract}
\section{Introduction}

\textbf{Entity Linking (EL)} identifies mentions in text and links them to entities in a \textbf{Knowledge Graph (KG)}. Most EL research targets \textbf{general domain EL}, where the central challenge is \textbf{disambiguation} among candidates with similar surface forms (e.g., "Apple" the company vs.\ the fruit). \textbf{Zero-Shot Entity Linking (ZSEL)} aims to perform linking without requiring human-annotated training data. Recent ZSEL approaches \cite{wu2020scalable, xu2023read,zhou2024gendecider} have made significant progress by leveraging entity descriptions and contextual embeddings to resolve such ambiguities. However, these methods are typically trained on open-domain corpora such as Wikipedia, where mentions and entities often exhibit high lexical overlap.

\textbf{Scientific domain EL} faces a different challenge from general domain EL, where mentions and entity names often \textbf{lack lexical overlap}. Rather than resolving ambiguity among lexically similar candidates, scientific EL must handle highly variable terminology, where formal names may differ significantly from their common or colloquial names. For example, a paper may use ``lambing potential'' instead of ``goat fertility'', and ``hypertension'' instead of ``high blood pressure''. This \textbf{lexical divergence} poses a major obstacle for standard approaches, which rely on token-level similarity. Another unique characteristic of scientific EL is that the KG is often represented as an ontology, with terms organized in hierarchical structures that show relationships between concepts. Such information is not sufficiently leveraged in existing ZSEL models trained on general domain corpora.

Consequently, pretrained ZSEL models struggle with scientific EL tasks. First, due to differences in lexical challenges between general and specialized scientific domains, they struggle to transfer effectively.
Second, they cannot handle specialized terminology that the general domain rarely encounters.
Third, pretrained ZSEL models ignore the ontology structure, treating each entity as an isolated string and missing the hierarchical signal that distinguishes closely related concepts. All limitations point to fine-tuning on in-domain data as the natural remedy, and prior work has taken this route \cite{yuan2022generative,xu2023improving}. However, fine-tuning requires labeled mention-entity pairs, and expert annotation is expensive and unavailable in many specialized domains. Recent work uses LLMs to synthesize training data \cite{xin2025llmael,ye2025llm}. Still, these methods use an LLM for each mention: the token cost scales with the size of the corpus, and the generated pairs are tied to that corpus rather than being reusable across new corpora drawn from the same ontology.

To address these challenges, we propose \our, a cost-aware framework that does not require human-labeled mention-entity pairs from the target domain. The framework generates pseudo-labeled training pairs by prompting an LLM for alias names that serve as lexical bridges between corpus mentions and ontology entities. To bound LLM cost, we apply this augmentation on the entity side rather than the mention side: instead of issuing one LLM call per corpus mention, we identify a small set of entities and generate aliases for each, then use those aliases to find mentions to form mention-entity pairs for fine-tuning. To ensure that LLM-generated aliases properly align with the entities in the ontology, we further apply an ontology-aware filter that uses entity hierarchy and semantic similarity to discard drifted aliases before training.

We further release a new EL benchmark drawn from animal science literature. This dataset is curated by domain experts and highlights the lexical divergence challenge in scientific EL tasks. 

Our main contributions are as follows:
\begin{itemize}
    \item A new animal science EL benchmark dataset formed based on PubMed articles linked to three livestock trait ontologies.
    
    \item \textbf{\our}, a zero-shot EL framework that uses an LLM to generate alias names for selected entities and builds pseudo pairs from real mentions and original entity names.
    
    \item \our improves over the non-fine-tuned baseline across all five benchmarks; moreover, combining \our generated aliases with curated synonyms achieves the best performance in most settings.
\end{itemize}

\section{Related Work}\label{sec:relatedWork}

General domain EL primarily targets \emph{disambiguation} among lexically similar candidates. Early systems exploited surface overlap and context heuristics~\cite{cucerzan2007large,ratinov2011local}, and later methods used pretrained language models and introduced dense semantic representations that improved robustness to surface variation~\cite{yamada2016joint,gillick2019learning} where they assume substantial lexical overlap between mention and entity, which is an assumption that fails in scientific text. 

\subsection{Zero-Shot Entity Linking}
ZSEL removes dependence on in-domain labelled mentions by linking through entity descriptions that enable generalization to unseen entities~\cite{logeswaran2019zero,wu2020scalable,xu2023read,zhou2024gendecider}. Existing ZSEL methods show strong performance on Wikipedia-style benchmarks where mentions and entity names have lexical overlap, but they implicitly rely on surface-form similarity and transfer poorly to specialized domains where mentions and entities may share no lexical overlap at all. Generative EL systems, such as GENRE~\citep{decao2021autoregressive}, struggle with unseen entities in the zero-shot setting.

\subsection{Biomedical Entity Linking}
Biomedical EL faces this lexical divergence most acutely: abbreviations, acronyms, and synonyms produce mention-entity pairs with minimal surface overlap (e.g., ``EGFR'' vs.\ ``Epidermal Growth Factor Receptor''). Early dictionary and rule-based methods offer high precision but limited scalability~\cite{aronson2001effective,leaman2013dnorm}. Neural approaches with domain-specific encoders~\cite{sung2020biomedical, liu2021self,chen2021lightweight}, generative formulations~\cite{yuan2022generative}, and cross-entity interaction~\cite{xu2023improving,kim2025learning} can improve performance but still depend on large expert-annotated corpora for fine-tuning, which are unavailable in emerging scientific domains where ontology synonyms are also sparse.

\subsection{LLM-Based Data Augmentation}
Recent work uses LLMs to synthesize EL supervision, either by augmenting mention contexts or by acting as end-to-end disambiguators~\cite{xin2025llmael,sanz2025accelerating,ye2025llm}. These approaches generate per-mention, so token cost scales with corpus size, and there is no guard against drift. \our inverts the direction: it generates aliases for a subset of entities, pairs them with observed corpus mentions, and applies a filter to discard drifted aliases, which combines the semantic reach of LLMs with the efficiency of a compact retriever-reranker pipeline.
\section{Animal Science Benchmark}
\label{sec:benchmark}

Animal science research on quantitative trait loci (QTL) requires normalizing free-text trait mentions to enable combining findings across studies. Trait names, however, differ across research aspects, species, and product lines, and are therefore curated in separate ontologies. 

In the proposed animal science benchmark, the EL task utilizes three livestock trait ontologies, Clinical Measurement Ontology (CMO), Vertebrate Trait Ontology (VT), and Livestock Product Trait Ontology (LPT), because they are used widely for animal QTL studies. The EL task expects recognised entities to perform linking, and in order to follow the zero-shot setting, we do not annotate a training set. Instead, we use the AnimalQTLdb Benchmark released in~\citet{hu2007animalqtldb} as the training corpus and use the CuPUL method ~\citep{li2025examine} to recognize trait mentions.

The test dataset is annotated manually using two strategies. First, a domain expert randomly selected 150 records from AnimalQTLdb, an animal science database curated for QTL research. Each record contains a normalized trait name, the ontology IDs from the aforementioned 3 ontologies, and the corresponding publication (PubMed ID) from which the record is curated. The expert then manually identifies corresponding mentions from the full paper. Second, we obtain a comprehensive list of trait names and their commonly observed mentions from AnimalQTLdb curators. Then, we collect all abstracts of papers curated in AnimalQTLdb and apply string matching to identify the trait mentions. Four student annotators manually identified the entities from the three ontologies and corrected mention boundaries when needed. Finally, the domain expert validated the annotations.

\begin{table*}[t]
\centering
\small
\setlength{\tabcolsep}{4pt}
\renewcommand{\arraystretch}{1.01}
\begin{tabular}{ll|cccccc|cccc}
\hline
 & & \multicolumn{6}{c|}{Test set statistics} & \multicolumn{4}{c}{Ontology statistics} \\
\cline{3-8} \cline{9-12}
Dataset name & Ontology Name & Samples & HO\% & LO\% & NO\% & Ment & Ent & Entities & Synonyms & Syn/Ent & Cov\% \\
\hline\hline
NCBI Disease & MEDIC & 960  & 32.08 & 45.00 & 22.92 & 287  & 190  & 13{,}316  & 127{,}370 & 9.57 & 85.0 \\
BC5CDR       & MeSH  & 9465 & 62.07 & 13.77 & 24.16 & 2084 & 1267 & 355{,}213 & 757{,}086 & 2.13 & 94.5 \\
QTL\textsubscript{CMO}        & CMO   & 2032 & 20.13 & 36.42 & 43.45 & 458  & 107  & 4{,}133   & 4{,}413   & 1.07 & 57.2 \\
QTL\textsubscript{VT}            & VT    & 1688 & 8.83  & 34.06 & 57.11 & 500  & 111  & 4{,}044   & 3{,}670   & 0.91 & 44.0 \\
QTL\textsubscript{LPT}           & LPT   & 722  & 30.75 & 24.24 & 45.01 & 233  & 103  & 520       & 462       & 0.89 & 41.7 \\
\hline
\end{tabular}
\caption{Combined test-set and ontology statistics. Test set block: Samples refers to the number of mention-entity pairs in the test set, HO, LO, and NO show the percentages of samples by overlap categories; and Ment and Ent refer to the number of unique mentions and unique entities in the test set, respectively. Ontology block: Entities and Synonyms refers to the total number in the ontology, Syn/Ent refers to averaged synonyms per entity, and Cov\% is the percentage of entities with at least one synonym provided. Note that QTL\textsubscript{CMO}, QTL\textsubscript{VT}, and QTL\textsubscript{LPT} are test-only datasets.}
\label{tab:dataset-stats}
\end{table*}

The statistics of two existing biomedical benchmarks NCBI Disease~\cite{dougan2014ncbi}, BC5CDR~\cite{li2016biocreative}, and the new animal science benchmarks (QTL\textsubscript{CMO}, QTL\textsubscript{VT}, and QTL\textsubscript{LPT}) are summarized in
Table~\ref{tab:dataset-stats}.

To characterize different levels of lexical divergence, we partition mention--entity pairs into three categories based on their degree of lexical overlap. \textbf{HO} (high overlap) refers to cases where the mention is a substring of the entity name. \textbf{LO} (low overlap) refers to cases where the mention and entity name exhibit partial lexical overlap, but the mention is not a substring of the entity name. \textbf{NO} (no overlap) refers to cases where the mention and entity name share no lexical overlap.

The resulting test sets differ from existing biomedical benchmarks in two ways. First, the NO rate is substantially higher than existing benchmarks. Moreover, the high ratio of unique mentions to unique entities indicates that the mentions are highly diverse for the same entities.
On the other hand, the three ontologies from the animal science domain are less developed than those in the biomedical domains, with a lower percentage of entities with synonyms and fewer synonyms per entity. Both observations illustrate that our new animal science benchmark may be significantly harder than the existing biomedical benchmarks.

\textbf{Annotation Limitations:} Due to the high cost of domain-expert review, the test sets are small, which limits statistical power for fine-grained comparisons. The pipeline may also carry a selection bias: we select mentions using dictionary string matching against a list of commonly observed mentions provided by domain experts, so mentions absent from those lists are dropped before annotators see them. 
However, the list is significantly larger than synonyms provided in ontologies and contains more diverse expressions. For example, ``IMF'' in this list is widely used to refer to "intramuscular fat" in QTL articles but does not appear in any ontologies. Further, annotators partially offset this by manually collecting no-overlap cases. However, the test mention pool remains biased toward surface forms that the ontologies anticipate.
\section{Preliminaries}
\label{sec:preliminaries}

Let $\mathcal{E}$ be the set of entities in a target ontology and $\mathcal{M}$ the mentions extracted from an unlabeled training corpus. Each entity $e \in \mathcal{E}$ has a name $n(e)$, a definition $d(e)$, and may have a list of synonyms $\mathrm{Syn}(e)$. Given a mention $m \in \mathcal{M}$ with context $c(m)$, the task is to predict its referent entity $e \in \mathcal{E}$. For the zero-shot setting, no human-labeled $\langle m, e \rangle$ pairs are available for training. In this paper, we use \emph{zero-shot} to denote the zero-human-label setting.

A bi-encoder retriever maps the mention and the entity to dense vectors $\tau_m, \tau_e \in \mathbb{R}^d$ and scores them by their dot product $\tau_m \cdot \tau_e$. It returns the top-$K$ entities under this score. We use the pretrained BLINK~\cite{wu2020scalable} as the bi-encoder backbone; details in Appendix~\ref{sec:biencoder-appendix}.

A cross-encoder reranker encodes mention $m$, its context $c(m)$, and all candidates $e$ jointly to produce a score $s(m, e)$. We consider two pretrained cross-encoder backbones, BLINK~\cite{wu2020scalable} and ReS~\cite{xu2023read}; details in Appendix~\ref{sec:reranker-appendix}.
\section{Methodology}
\label{sec:methodology}

\begin{figure*}[t]
\centering
\includegraphics[width=\textwidth]{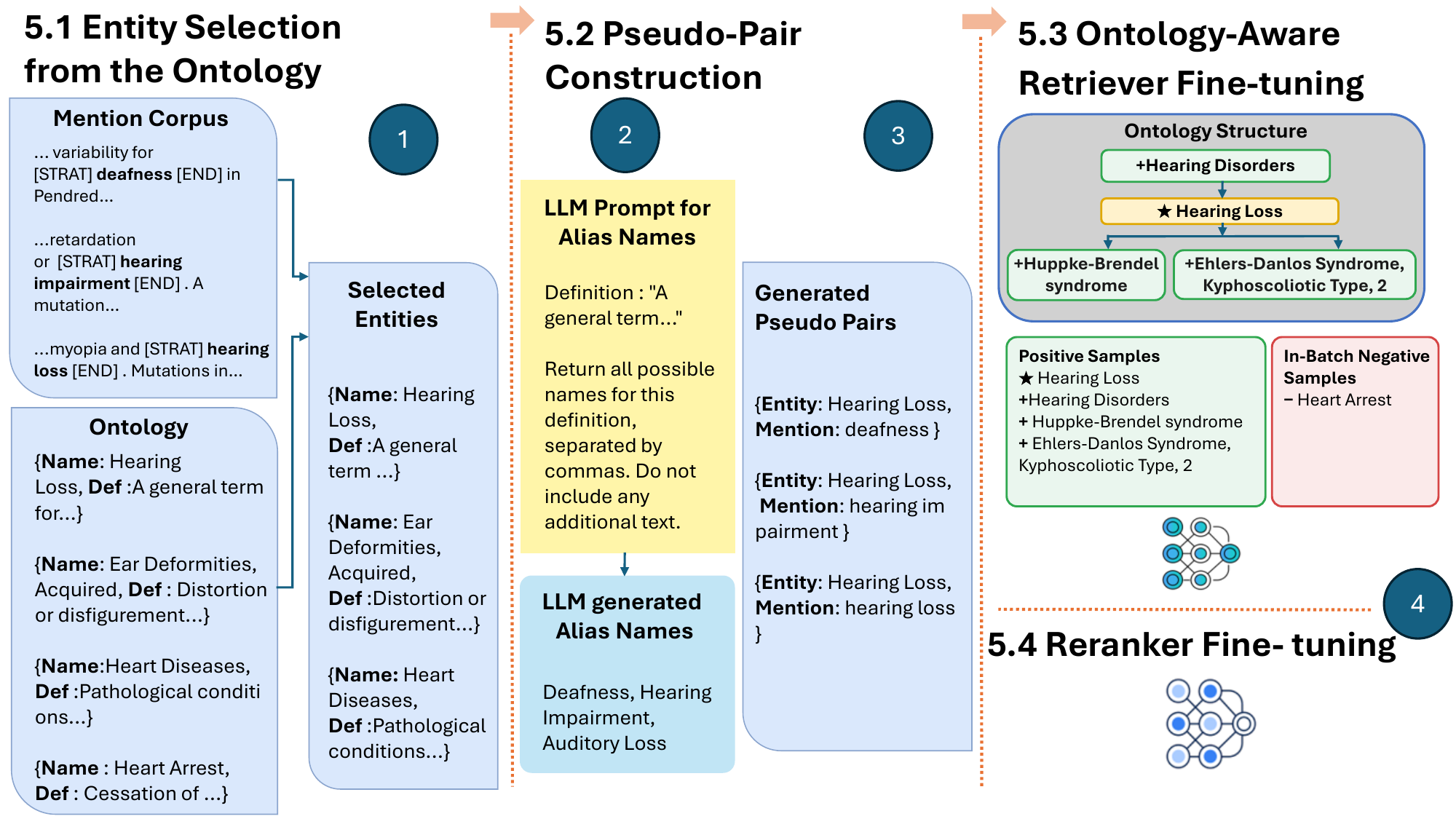}
\caption{Overview of the \our framework with a worked example. \textbf{(1)} Candidate entities are selected from the ontology using the unlabeled mention corpus. \textbf{(2)} Selected entity definitions are sent to the LLM to generate aliases. \textbf{(3)} The pseudo-pairs are used to fine-tune the retriever, where ontology neighbors of the gold entity (parents and children) are promoted as positives ($\bigstar$ gold entity, $+$ positive, $-$ in-batch negative) and other entities within the batch act as in-batch negatives. \textbf{(4)} The pseudo-pairs are used to fine-tune the reranker.}
\label{fig:scizsel-overview}
\end{figure*}

\subsection{Entity Selection from the Ontology}
\label{sec:entity-selection}

An ontology defines the vocabulary of a domain and often contains a large collection of entities. However, following the assumption of Zipf’s law~\citep{zipf1949human}, many of those entities are not mentioned in the target corpus, so it is unnecessary to consider every entity. Therefore, to reduce cost, Sci-ZSEL strategically selects a subset of entities and uses LLM to generate their aliases. This also makes the generated supervision reusable across corpora collected from similar domains.
Figure~\ref {fig:scizsel-overview} shows the entire flow of pseudo pair generation.

In scientific domains, mention ambiguity is relatively low. If a mention matches the surface name of an entity, the pairing is usually correct. Following this assumption, we include these entities whose name appears as mentions in the corpus:
\begin{equation}
\mathcal{E}_{\mathrm{EM}} = \big\{\, e \in \mathcal{E} : \exists\, m \in \mathcal{M},\; m \equiv n(e) \,\big\} \:.
\label{eq:eem}
\end{equation}

A pretrained BLINK bi-encoder embeds mentions and entities into a semantic space, so top-ranked entities can provide a semantic bridge for lexically divergent mentions. To do so, we pass each mention $m \in \mathcal{M}$ to the bi-encoder and use its top-1 entity. Specifically, we define this set as $\mathcal{E}_{\mathrm{BT}}$:

{
\begin{equation}
\mathcal{E}_{\mathrm{BT}} = \big\{\, \mathrm{top1}_{\mathrm{BLINK}}(m) : m \in \mathcal{M} \,\big\} \:.
\label{eq:ebt}
\end{equation}
}

\subsection{Pseudo-Pair Construction}
\label{sec:pseudo-pairs}
We construct pseudo-labeled pairs by using the unlabeled training corpus and the selected entities ($\mathcal{E}_{\mathrm{EM}} \cup \mathcal{E}_{\mathrm{BT}}$) from Section~\ref{sec:entity-selection} via the three construction strategies below.

\subsubsection{Construction from Surface Name Matching} 
Similar to exact match entity selection, but here, we return $\langle m, e \rangle$ pair for the pseudo training data instead of entity. Mathematically, it is defined as:
\begingroup
{
\small
\setlength{\abovedisplayskip}{6pt}
\setlength{\belowdisplayskip}{2pt}
\setlength{\abovedisplayshortskip}{6pt}
\setlength{\belowdisplayshortskip}{2pt}
\begin{equation}
\mathcal{P}_1 = \big\{\, \langle m, e \rangle : m \in \mathcal{M},\; e \in \mathcal{E},\; m \equiv n(e) \,\big\}.
\label{eq:p1}
\end{equation}
\par
}
\endgroup
\subsubsection{Construction from LLM-generated 
Alias}\label{sec:construction}
LLMs possess semantically related lexical expression capabilities; we leverage them to generate representative aliases from entity definitions. Our goal is to synthesize valid surface forms that mentions in the corpus might use. To do so, we prompt the LLM with the definition $d(e)$ of an entity $e \in \mathcal{E}_{\mathrm{EM}} \cup \mathcal{E}_{\mathrm{BT}}$ to produce aliases $a(e)$ (the prompt template is provided in Appendix \ref{sec:gen-prompt}). We apply alias generation to all entities in  $\mathcal{E}_{\mathrm{EM}} \cup \mathcal{E}_{\mathrm{BT}}$, producing the alias set $\mathcal{E}_{\mathrm{A}}$.

\paragraph{Ontology-Aware Filtering.} 
\label{sec:onto-hierarchy-filtering}
A common failure mode of LLM generation is semantic drift, where an alias inadvertently describes a hierarchical neighbor of $e$ rather than the entity itself. To mitigate this, we filter the generated aliases $a(e)$ by comparing their semantic similarity to the target entity name $n(e)$ against the entity's ontology neighbors. 

We first define the ontology neighborhood of entity $e$ using its parents, children, and siblings:

{

\begin{align}
\mathcal{N}(e) &= \operatorname{parents}(e) \cup \operatorname{children}(e) \:, \label{eq:N}\\
\mathcal{N}^{+}(e) &= \mathcal{N}(e) \cup \operatorname{siblings}(e) \:. \label{eq:Nplus}
\end{align}
\par}

For each generated alias $a(e) \in \mathcal{E}_{\mathrm{A}}$, we compute an anchor similarity to the entity name $n(e)$ and a drift similarity to its most closely related neighbor in $\mathcal{N}^+(e)$:

\begin{align}
s_{\text{anchor}} &= \operatorname{sim}\!\bigl(n(e),\, a(e)\bigr) \label{eq:anchor}, \\
s_{\text{drift}}  &= \max_{e_\mathcal{N} \in \mathcal{N}^{+}(e)} \operatorname{sim}\!\bigl(n(e_\mathcal{N}),\, a(e)\bigr).\label{eq:drift}
\end{align}

where $\operatorname{sim}(\cdot,\cdot)$ denotes the cosine similarity between BioLORD~\cite{biolord} embeddings of two entity names. 

We discard the alias $a(e)$ if $s_{\text{anchor}} < s_{\text{drift}}$ or $s_{\text{anchor}} < \tau$, where $\tau = 0.9$.

The first condition removes aliases that are semantically closer to a neighboring entity than to the target entity $e$, while the second removes aliases that diverge excessively from the target entity name $n(e)$. The resulting filtered alias set is denoted as $\mathcal{E}_{\mathrm{AF}}$.

\paragraph {Pseudo-Label Construction.} Finally, we search for the filtered aliases  $a(e) \in \mathcal{E}_{\mathrm{AF}}$  within the unlabeled corpus $\mathcal{M}$. We form a pseudo-labeled pair whenever an observed mention $m$ exactly matches a filtered alias $a(e)$ (denoted as $m \equiv a(e)$ after whitespace and case normalization). Specifically, we generate the following sets of pseudo-labeled pairs:

{
\setlength{\abovedisplayskip}{2pt}%
\setlength{\belowdisplayskip}{2pt}%
\small
\begin{equation}
\mathcal{P}_2 = \{ \langle m, e \rangle : e \in \mathcal{E}_{\text{AF}}, m \in \mathcal{M}, m \equiv a(e) \} \:.
\label{eq:p2}
\end{equation}
}
The combined set is:
\begin{equation}
\mathcal{P}_{\mathrm{Gen}} = \mathcal{P}_1 \cup \mathcal{P}_2 \:.
\label{eq:pGen}
\end{equation}
\subsubsection{Construction from Ontology Synonym} 

Many ontologies contain expert-curated synonym lists designed to capture alternative surface forms of an entity's name. Therefore, we can directly leverage these curated synonyms, denoted as $s \in \text{Syn}(e)$, as highly reliable lexical bridges. We construct a pseudo-labeled pair for an entity $e$ whenever one of its curated synonyms $s$ exactly matches an observed mention $m$ in the unlabeled corpus. Formally, it is defined as:
\begin{equation}
\mathcal{P}_{\mathrm{Syn}} = \big\{\, \langle m, e \rangle : e \in \mathcal{E},\;\; m \in \mathcal{M},\; m \equiv s \,\big\} \:.
\label{eq:pSyn}
\end{equation}

\subsection{Ontology-Aware Retriever Fine-tuning}
\label{sec:biencoder-finetune}
The retriever determines which candidate entities are exposed to the reranker. We therefore fine-tune the retriever to better rank semantically related ontology entities. Hierarchical neighbors in an ontology often have related meanings; they provide informative training signals for improving retrieval under the ontology structure. To do that, we incorporate ontology-aware sampling strategies to the retriever. The \textbf{BASE} strategy treats every non-gold entity as a negative. The \textbf{RM-PCS} removes neighbors $\mathcal{N}^{+}(e)$ from the negatives. The \textbf{PC-POS} promotes parents and children $\mathcal{N}(e)$ of $e$, as they are added as positives, while siblings remain as negatives.

Each strategy uses an in-batch cross-entropy objective with these introduced positive and negative sets; the goal is to score positive pairs higher than negative pairs within each batch. Full formulation is in Appendix~\ref{sec:biencoder-neg-appendix}.

\subsection{Reranker Fine-tuning}
\label{sec:crossencoder-finetune}
The primary responsibility of the reranker is to learn the accurate distinctions among relevant top-$K$ candidates, then assign the highest score to the most accurate candidate. Therefore, we do not apply the ontology-aware sampling strategies here. Instead, we fine-tune the BLINK cross-encoder and ReS using their standard architecture (Section~\ref{sec:preliminaries}).

For each pseudo pair $\langle m, e \rangle$, we retrieve the top-$K$ candidates from the bi-encoder, treat $e$ as the positive and randomly selected candidates as negatives from the rest, then optimize a cross-entropy loss over the candidate set. The goal is to rank the pseudo-labeled entity at the top. Full formulation and the candidate-set augmentation rule (when $e$ is missing from top-$K$) are in Appendix~\ref{sec:reranker-appendix}.
\section{Experimental Studies}\label{sec:experiment}

\subsection{Datasets}
\label{sec:datasets}

We evaluate \our on five benchmarks. NCBI Disease~\cite{dougan2014ncbi} (linking to MEDIC~\cite{davis2012medic}) and BC5CDR~\cite{li2016biocreative} (linking to MeSH~\cite{lipscomb2000medical}) are existing synonym-rich biomedical benchmarks; QTL\textsubscript{CMO}, QTL\textsubscript{VT}, and QTL\textsubscript{LPT} are synonym-sparse animal science benchmarks released with this paper (Section~\ref{sec:benchmark}). Training pairs are constructed as described in Section~\ref{sec:pseudo-pairs}. Statistics for test sets and training pairs are in Table~\ref{tab:dataset-stats} and Table~\ref{tab:smpl_per_setting}, respectively.

\subsection{Baselines}
\label{sec:baselines}
We select baseline backbones that satisfy two requirements. First, a baseline must be a zero-shot method or otherwise applicable to an out-of-domain setting. Second, its pretrained model must be publicly accessible so that it can be directly used as a backbone in our framework.

Therefore, we instantiate \our on two pretrained general-domain ZSEL backbones: BLINK~\cite{wu2020scalable} for the retriever and reranker, and ReS~\cite{xu2023read} for the reranker. For each backbone on each dataset, we run five fine-tuning settings to isolate the contributions. The \textbf{No fine-tune} setting uses the frozen, pretrained ZSEL backbone to exhibit the performance under scientific domain shift. \textbf{\our w/o filter} fine-tunes on the unfiltered LLM-generated alias setting (omitting the step from Section ~\ref{sec:onto-hierarchy-filtering}) to see the raw impact.
 \textbf{\our} fine-tunes after filtering, so it uses $\mathcal{P}_{\text{Gen}}$ (Eq. ~\ref{eq:pGen}), which demonstrates the impact and necessity of the ontology-aware filter. \textbf{Synonym} uses $\mathcal{P}_{\mathrm{Syn}}$ (Eq.~\ref{eq:pSyn}) that exhibits whether curated synonyms alone suffice for EL, where they are abundant, and \textbf{\our + Synonym} $\mathcal{P}_{\mathrm{Gen}} \cup \mathcal{P}_{\mathrm{Syn}}$ shows whether LLM aliases and curated synonyms are complementary.

\subsection{Experimental Setup}
\label{sec:setups}
We fine-tune two pretrained zero-shot EL backbones, BLINK and ReS, using their original architectures. In a zero-shot setting, we do not have a validation set, so best-epoch selection is not possible; we therefore fix the number of epochs and report the results from the last epoch. All reported results are means over three random seeds (0, 42, 52313).
Details in Appendix~\ref{app:hyperparameters}.

\subsection{Evaluation Metrics}
\label{sec:metrics}

We evaluate the retriever using Recall@64 and the reranker using Recall@1. Recall@$K$ measures the proportion of mentions for which the gold entity appears among the top-$K$ candidates. We further report mean reciprocal rank (MRR), which measures the average reciprocal rank of the gold entity across test mentions, as well as results on the HO, LO, and NO subsets defined in Section~\ref{sec:benchmark} to assess performance under varying levels of lexical divergence.

\subsection{Experimental Results}
In this section, we begin by analyzing the quality of our pseudo training pairs in Section ~\ref{sec:pseudo-quality}. Next, we evaluate the retriever's performance, presenting its overall Recall@64 scores in Table ~\ref{tab:retriever_recall64} and providing MRR and category-wise performance (HO, LO, and NO) in Tables ~\ref{tab:recall64_base}, ~\ref{tab:recall64_rmpcs}, and ~\ref{tab:recall64_pcpos}. Then, we evaluate the reranker's performance in Table~\ref{tab:reranker_recall1}, highlighting its overall Recall@1, MRR, and category-wise results. To ensure our findings are stable and reliable, every number we report is an average calculated across three separate test runs using random seeds 0, 42, and 52313.

\begin{table}[t]
\centering
\small
\setlength{\tabcolsep}{2pt}
\renewcommand{\arraystretch}{1.1}
\begin{tabular}{l|ccc|ccc}
\hline
 & \multicolumn{3}{c|}{Exact match ($\mathcal{E}_\text{EM}$) } & \multicolumn{3}{c}{Biencoder top-1 ($\mathcal{E}_\text{BT}$)} \\
\cline{2-4} \cline{5-7}
Ontology & Ent & \%Syn & \%Name & Ent & \%Syn & \%Name \\
\hline\hline
MEDIC & 169 & 26.04 & 54.44 & 636  & 29.87 & 29.09 \\
MeSH  & 721 & 10.54 & 53.68 & 1341 & 8.58  & 41.83 \\
CMO   & 58  & 5.17  & 36.21 & 1260 & 3.33  & 10.00 \\
VT    & 20  & 0.00  & 40.00 & 1396 & 2.22  & 7.88  \\
LPT   & 30  & 3.33  & 46.67 & 361  & 2.22  & 14.68 \\
\hline
\end{tabular}
\caption{LLM-generated alias behavior across source ontologies. Ent is the number of entities used for alias generation; \%Syn and \%Name are the percentages of aliases matching a curated synonym and the original entity name, respectively (the remainder are new lexical bridges).}
\label{tab:llm-alias}
\end{table}

\begin{table}[h]
\setlength{\tabcolsep}{1.5pt}
\renewcommand{\arraystretch}{1.05}
\centering
\small
\begin{tabular}{llrr}
\toprule
Dataset&Setting&Accuracy& pseudo pair count\\
\midrule
NCBI   & $P_{\text{Gen}}$ W/O filter & 72.91 & 1757\\
       & \;$\mathcal{P}_{\text{Gen}}$       & \textbf{90.32} & 1301\\
\midrule
BC5CDR & $P_{\text{Gen}}$ W/O filter & 72.55 & 7447\\
       & \; $\mathcal{P}_{\text{Gen}}$       & \textbf{91.4} & 5766\\
\bottomrule
\end{tabular}
\caption{Pseudo pair accuracy and number of generated pairs, before and after applying ontology-aware filtering on the NCBI and BC5CDR training sets.}
\label{tab:filter-acc}
\end{table}

\begin{table}[t]
\centering
\setlength{\tabcolsep}{4pt}
\renewcommand{\arraystretch}{1.05}
\small
\resizebox{0.5\textwidth}{!}{%
\begin{tabular}{|c|l|c|c|c|}
\hline
\textbf{Dataset} & \textbf{Setting} & \textbf{BASE} & \textbf{RM-PCS} & \textbf{PC-POS} \\
\hline
\hline
\multirow{5}{*}{\textbf{NCBI}} & No fine-tune & 88.12$_{\pm 0.00}$ & 88.12$_{\pm 0.00}$ & 88.12$_{\pm 0.00}$ \\
 & \our w/o filter & 87.74$_{\pm 0.22}$ & 87.01$_{\pm 1.85}$ & \textbf{88.75$_{\pm 0.36}$} \\
 & \our & 87.40$_{\pm 0.47}$ & 80.70$_{\pm 11.97}$ & \textbf{87.81$_{\pm 0.63}$} \\
 & Synonym & 89.20$_{\pm 0.24}$ & 89.73$_{\pm 0.30}$ & \textbf{90.21$_{\pm 0.36}$} \\
 & \our+ Synonym & 89.72$_{\pm 0.97}$ & 89.41$_{\pm 0.78}$ & \textbf{90.80$_{\pm 0.87}$} \\
\hline
\multirow{5}{*}{\textbf{BC5CDR}} & No fine-tune & 88.79$_{\pm 0.00}$ & 88.79$_{\pm 0.00}$ & 88.79$_{\pm 0.00}$ \\
 & \our w/o filter & 88.41$_{\pm 0.74}$ & 88.21$_{\pm 2.01}$ & \textbf{88.93$_{\pm 1.03}$} \\
 & \our & 85.27$_{\pm 1.75}$ & 85.41$_{\pm 0.27}$ & \textbf{87.80$_{\pm 0.74}$} \\
 & Synonym & 87.22$_{\pm 0.16}$ & 87.13$_{\pm 0.59}$ & \textbf{88.57$_{\pm 0.50}$} \\
 & \our+ Synonym & 87.96$_{\pm 0.48}$ & 86.61$_{\pm 2.16}$ & \textbf{89.14$_{\pm 0.65}$} \\
\hline
\multirow{5}{*}{\textbf{QTL\textsubscript{CMO} }} & No fine-tune & 75.74$_{\pm 0.00}$ & 75.74$_{\pm 0.00}$ & 75.74$_{\pm 0.00}$ \\
 & \our w/o filter & \textbf{89.96$_{\pm 0.23}$} & 88.02$_{\pm 1.24}$ & 89.24$_{\pm 1.61}$ \\
 & \our & \textbf{89.22$_{\pm 0.86}$} & 88.06$_{\pm 0.25}$ & 88.34$_{\pm 1.37}$ \\
 & Synonym & \textbf{85.71$_{\pm 0.40}$} & 85.47$_{\pm 0.49}$ & 85.45$_{\pm 0.91}$ \\
 & \our+ Synonym & 88.22$_{\pm 0.89}$ & 89.03$_{\pm 1.68}$ & \textbf{89.09$_{\pm 1.57}$} \\
\hline
\multirow{5}{*}{\textbf{QTL\textsubscript{VT} }} & No fine-tune & 72.63$_{\pm 0.00}$ & 72.63$_{\pm 0.00}$ & 72.63$_{\pm 0.00}$ \\
 & \our w/o filter & \textbf{84.70$_{\pm 0.96}$} & 84.52$_{\pm 1.08}$ & 83.69$_{\pm 0.94}$ \\
 & \our & 84.72$_{\pm 1.69}$ & \textbf{85.51$_{\pm 0.30}$} & 85.31$_{\pm 1.03}$ \\
 & Synonym & 86.99$_{\pm 1.74}$ & 86.79$_{\pm 2.67}$ & \textbf{87.91$_{\pm 2.22}$} \\
 & \our+ Synonym & 85.58$_{\pm 4.10}$ & \textbf{88.69$_{\pm 2.22}$} & 87.99$_{\pm 1.47}$ \\
\hline
\multirow{5}{*}{\textbf{QTL\textsubscript{LPT} }} & No fine-tune & 68.70$_{\pm 0.00}$ & 68.70$_{\pm 0.00}$ & 68.70$_{\pm 0.00}$ \\
 & \our w/o filter & 70.08$_{\pm 0.28}$ & 70.82$_{\pm 0.97}$ & \textbf{71.01$_{\pm 0.92}$} \\
 & \our & 75.07$_{\pm 1.54}$ & \textbf{76.87$_{\pm 0.73}$} & 76.18$_{\pm 0.37}$ \\
 & Synonym & \textbf{92.43$_{\pm 0.68}$} & 91.83$_{\pm 0.42}$ & 91.78$_{\pm 1.24}$ \\
 & \our+ Synonym & 92.29$_{\pm 0.44}$ & 92.48$_{\pm 0.35}$ & \textbf{92.66$_{\pm 0.69}$} \\
\hline
\end{tabular}
}
\caption{A comparison of retriever Recall@64 across benchmarks under various negative-sampling configurations. \textbf{Bold} indicates the best results when comparing column-wise.}
\label{tab:retriever_recall64}
\end{table}

\begin{table*}[t]
\centering
\setlength{\tabcolsep}{3pt}
\renewcommand{\arraystretch}{1.05}

\scriptsize 
\resizebox{\textwidth}{!}{%

\begin{tabular}{|c|l|c|c|c|c|c|c|c|c|c|c|}

\hline
 &  & \multicolumn{5}{c|}{\textbf{BLINK}} & \multicolumn{5}{c|}{\textbf{ReS}} \\
\cline{3-12}
\textbf{Dataset} & \textbf{Setting} & \textbf{Overall} & \textbf{MRR} & \textbf{HO} & \textbf{LO} & \textbf{NO} & \textbf{Overall} & \textbf{MRR} & \textbf{HO} & \textbf{LO} & \textbf{NO} \\
\hline
\hline
\multirow{5}{*}{\textbf{NCBI}} & No fine-tune & 64.38$_{\pm 0.00}$ & 72.39$_{\pm 0.00}$ & \textbf{93.83$_{\pm 0.00}$} & 64.12$_{\pm 0.00}$ & 23.64$_{\pm 0.00}$ & 45.94$_{\pm 0.00}$ & 58.41$_{\pm 0.00}$ & 67.53$_{\pm 0.00}$ & 44.44$_{\pm 0.00}$ & 18.64$_{\pm 0.00}$ \\
 & \our w/o filter & 70.69$_{\pm 0.58}$ & 77.65$_{\pm 0.47}$ & 92.75$_{\pm 0.19}$ & 66.44$_{\pm 1.22}$ & 48.18$_{\pm 0.79}$ & 60.28$_{\pm 0.42}$ & 73.57$_{\pm 0.27}$ & 68.72$_{\pm 0.68}$ & 59.41$_{\pm 1.88}$ & 50.15$_{\pm 3.19}$ \\
 & \our & 71.88$_{\pm 0.42}$ & 78.48$_{\pm 0.35}$ & 92.86$_{\pm 0.00}$ & 68.90$_{\pm 1.05}$ & 48.33$_{\pm 0.27}$ & 63.06$_{\pm 2.25}$ & 73.00$_{\pm 2.50}$ & 73.05$_{\pm 1.17}$ & 68.52$_{\pm 2.34}$ & 38.33$_{\pm 6.45}$ \\
 & Synonym & 76.18$_{\pm 0.24}$ & 83.17$_{\pm 0.49}$ & 93.18$_{\pm 0.33}$ & 73.23$_{\pm 0.58}$ & 58.18$_{\pm 2.08}$ & \textbf{78.09$_{\pm 0.73}$} & \textbf{84.59$_{\pm 0.37}$} & \textbf{93.94$_{\pm 2.40}$} & \textbf{74.07$_{\pm 0.70}$} & \textbf{63.79$_{\pm 1.89}$} \\
 & \our+ Synonym & \textbf{77.43$_{\pm 0.67}$} & \textbf{84.35$_{\pm 0.74}$} & 92.75$_{\pm 1.50}$ & \textbf{74.54$_{\pm 2.06}$} & \textbf{61.67$_{\pm 1.15}$} & 70.52$_{\pm 1.02}$ & 79.11$_{\pm 1.58}$ & 74.24$_{\pm 1.46}$ & 71.30$_{\pm 0.24}$ & \textbf{63.79$_{\pm 3.09}$} \\
\hline
\multirow{5}{*}{\textbf{BC5CDR}} & No fine-tune & 74.36$_{\pm 0.00}$ & 79.53$_{\pm 0.00}$ & \textbf{96.26$_{\pm 0.00}$} & 66.77$_{\pm 0.00}$ & 22.43$_{\pm 0.00}$ & 32.65$_{\pm 0.00}$ & 45.34$_{\pm 0.00}$ & 39.86$_{\pm 0.00}$ & 35.61$_{\pm 0.00}$ & 12.42$_{\pm 0.00}$ \\
 & \our w/o filter & 78.14$_{\pm 0.47}$ & 83.90$_{\pm 0.36}$ & 94.94$_{\pm 0.64}$ & 61.55$_{\pm 1.31}$ & 44.45$_{\pm 1.04}$ & 68.14$_{\pm 0.76}$ & 77.01$_{\pm 0.67}$ & 82.77$_{\pm 0.53}$ & 51.04$_{\pm 2.87}$ & 40.31$_{\pm 1.32}$ \\
 & \our & 79.16$_{\pm 0.58}$ & 84.40$_{\pm 0.27}$ & 96.01$_{\pm 0.31}$ & 72.17$_{\pm 2.03}$ & 39.88$_{\pm 1.45}$ & 76.91$_{\pm 0.53}$ & 82.52$_{\pm 0.28}$ & \textbf{94.83$_{\pm 0.58}$} & 64.65$_{\pm 1.58}$ & 37.85$_{\pm 1.42}$ \\
 & Synonym & 78.20$_{\pm 1.24}$ & 83.49$_{\pm 1.88}$ & 96.17$_{\pm 0.24}$ & 67.97$_{\pm 1.93}$ & 37.85$_{\pm 3.51}$ & 74.74$_{\pm 0.40}$ & 80.15$_{\pm 0.48}$ & 94.12$_{\pm 0.44}$ & 58.68$_{\pm 3.04}$ & 34.09$_{\pm 0.83}$ \\
 & \our+ Synonym & \textbf{81.20$_{\pm 0.58}$} & \textbf{86.20$_{\pm 0.48}$} & 95.96$_{\pm 0.30}$ & \textbf{74.93$_{\pm 1.96}$} & \textbf{46.87$_{\pm 1.30}$} & \textbf{77.90$_{\pm 0.12}$} & \textbf{83.21$_{\pm 0.26}$} & 94.78$_{\pm 0.17}$ & \textbf{65.42$_{\pm 2.49}$} & \textbf{41.66$_{\pm 0.66}$} \\
\hline
\multirow{5}{*}{\textbf{QTL\textsubscript{CMO} }} & No fine-tune & 50.15$_{\pm 0.00}$ & 60.74$_{\pm 0.00}$ & 92.91$_{\pm 0.00}$ & 50.95$_{\pm 0.00}$ & 29.67$_{\pm 0.00}$ & 48.92$_{\pm 0.00}$ & 61.06$_{\pm 0.00}$ & 82.15$_{\pm 0.00}$ & 51.49$_{\pm 0.00}$ & 31.37$_{\pm 0.00}$ \\
 & \our w/o filter & 57.09$_{\pm 2.37}$ & 66.43$_{\pm 1.96}$ & 93.97$_{\pm 3.88}$ & 58.74$_{\pm 2.26}$ & 38.62$_{\pm 2.31}$ & 41.42$_{\pm 2.31}$ & 54.82$_{\pm 2.24}$ & 44.34$_{\pm 2.98}$ & 49.59$_{\pm 4.06}$ & 33.22$_{\pm 4.05}$ \\
 & \our & 56.89$_{\pm 0.48}$ & 67.15$_{\pm 0.34}$ & 93.16$_{\pm 0.42}$ & 56.89$_{\pm 0.14}$ & \textbf{40.09$_{\pm 1.08}$} & 56.84$_{\pm 1.46}$ & 66.72$_{\pm 1.46}$ & 96.74$_{\pm 0.79}$ & 58.43$_{\pm 1.97}$ & \textbf{37.03$_{\pm 2.98}$} \\
 & Synonym & 57.59$_{\pm 1.72}$ & 66.50$_{\pm 1.51}$ & 93.15$_{\pm 0.25}$ & 62.79$_{\pm 2.24}$ & 36.77$_{\pm 2.34}$ & \textbf{58.48$_{\pm 0.37}$} & \textbf{68.29$_{\pm 0.36}$} & 94.70$_{\pm 0.51}$ & 64.91$_{\pm 0.41}$ & 36.32$_{\pm 1.09}$ \\
 & \our+ Synonym & \textbf{58.88$_{\pm 3.30}$} & \textbf{67.72$_{\pm 2.92}$} & \textbf{94.87$_{\pm 0.65}$} & \textbf{65.22$_{\pm 2.99}$} & 36.88$_{\pm 5.58}$ & 57.43$_{\pm 0.23}$ & 66.95$_{\pm 0.12}$ & \textbf{96.99$_{\pm 0.79}$} & \textbf{65.54$_{\pm 0.36}$} & 32.31$_{\pm 0.91}$ \\
\hline
\multirow{5}{*}{\textbf{QTL\textsubscript{VT} }} & No fine-tune & 43.13$_{\pm 0.00}$ & 56.38$_{\pm 0.00}$ & 97.32$_{\pm 0.00}$ & 48.70$_{\pm 0.00}$ & 31.43$_{\pm 0.00}$ & 38.33$_{\pm 0.00}$ & 51.14$_{\pm 0.00}$ & 95.97$_{\pm 0.00}$ & 54.61$_{\pm 0.00}$ & 19.71$_{\pm 0.00}$ \\
 & \our w/o filter & 38.29$_{\pm 1.21}$ & 54.51$_{\pm 0.74}$ & 97.32$_{\pm 0.00}$ & 50.90$_{\pm 2.58}$ & 21.64$_{\pm 0.94}$ & 45.81$_{\pm 2.13}$ & 59.78$_{\pm 0.86}$ & 97.32$_{\pm 0.00}$ & 64.12$_{\pm 5.84}$ & 26.94$_{\pm 0.40}$ \\
 & \our & 46.64$_{\pm 0.54}$ & 60.04$_{\pm 0.26}$ & 97.32$_{\pm 0.00}$ & 49.68$_{\pm 0.96}$ & 37.00$_{\pm 0.43}$ & 41.15$_{\pm 0.42}$ & 53.98$_{\pm 0.38}$ & 97.32$_{\pm 0.00}$ & 59.88$_{\pm 1.50}$ & 21.30$_{\pm 0.76}$ \\
 & Synonym & 70.83$_{\pm 0.42}$ & \textbf{79.66$_{\pm 0.73}$} & 97.54$_{\pm 0.39}$ & \textbf{76.64$_{\pm 1.40}$} & 63.24$_{\pm 1.32}$ & 62.78$_{\pm 3.91}$ & 73.65$_{\pm 3.55}$ & 96.87$_{\pm 0.39}$ & 67.54$_{\pm 0.61}$ & 54.67$_{\pm 6.58}$ \\
 & \our+ Synonym & \textbf{71.25$_{\pm 2.58}$} & 79.49$_{\pm 1.65}$ & \textbf{97.77$_{\pm 0.77}$} & 76.46$_{\pm 2.47}$ & \textbf{64.04$_{\pm 3.43}$} & \textbf{65.58$_{\pm 2.62}$} & \textbf{75.76$_{\pm 2.03}$} & \textbf{97.77$_{\pm 0.77}$} & \textbf{74.09$_{\pm 4.31}$} & \textbf{55.53$_{\pm 2.90}$} \\
\hline
\multirow{5}{*}{\textbf{QTL\textsubscript{LPT} }} & No fine-tune & 57.34$_{\pm 0.00}$ & 63.29$_{\pm 0.00}$ & 94.59$_{\pm 0.00}$ & 78.29$_{\pm 0.00}$ & 20.62$_{\pm 0.00}$ & 59.70$_{\pm 0.00}$ & 67.91$_{\pm 0.00}$ & 92.34$_{\pm 0.00}$ & 77.71$_{\pm 0.00}$ & 27.69$_{\pm 0.00}$ \\
 & \our w/o filter & 54.62$_{\pm 1.06}$ & 61.67$_{\pm 0.85}$ & 95.05$_{\pm 0.46}$ & 72.00$_{\pm 2.29}$ & 17.64$_{\pm 1.45}$ & 55.82$_{\pm 0.64}$ & 64.20$_{\pm 0.84}$ & 93.54$_{\pm 2.03}$ & 68.95$_{\pm 1.44}$ & 22.97$_{\pm 0.89}$ \\
 & \our & 58.22$_{\pm 0.29}$ & 64.18$_{\pm 0.19}$ & 95.35$_{\pm 0.52}$ & \textbf{79.81$_{\pm 1.43}$} & 21.23$_{\pm 0.62}$ & 62.14$_{\pm 0.81}$ & 69.71$_{\pm 0.60}$ & 94.29$_{\pm 0.26}$ & 80.38$_{\pm 1.32}$ & 30.36$_{\pm 0.99}$ \\
 & Synonym & 78.81$_{\pm 1.92}$ & 84.56$_{\pm 1.60}$ & 97.90$_{\pm 0.26}$ & 79.43$_{\pm 0.57}$ & 65.44$_{\pm 4.20}$ & 77.93$_{\pm 1.53}$ & 83.75$_{\pm 1.14}$ & 95.05$_{\pm 0.46}$ & 77.14$_{\pm 0.57}$ & 66.67$_{\pm 3.40}$ \\
 & \our+ Synonym & \textbf{80.70$_{\pm 0.21}$} & \textbf{86.12$_{\pm 0.12}$} & \textbf{98.05$_{\pm 0.26}$} & 79.05$_{\pm 3.15}$ & \textbf{69.74$_{\pm 1.28}$} & \textbf{81.21$_{\pm 2.29}$} & \textbf{86.21$_{\pm 1.38}$} & \textbf{97.45$_{\pm 0.52}$} & \textbf{80.95$_{\pm 1.19}$} & \textbf{70.26$_{\pm 4.48}$} \\
\hline
\end{tabular}%
}
\caption{ A performance comparison of BLINK and ReS rerankers across all five benchmarks. The Recall@1 reported as "Overall" where the score considers the entire dataset, while HO, LO, and NO show category-wise scores. \textbf{Bold} indicates the best results when comparing row-wise.}
\label{tab:reranker_recall1}
\end{table*}

\subsubsection{Cost-efficient Alias Generation for Lexical Diversity}
\label{sec:pseudo-quality}
\textbf{Cost-efficiency and Reusability.}
\our bounds LLM cost by querying aliases only for a selected subset of entities rather than every
corpus mention. Comparing the ontology size (Table~\ref{tab:dataset-stats}) against the selected set $\mathcal{E}_\text{EM} \cup \mathcal{E}_\text{BT}$ (Table~\ref{tab:llm-alias}), the efficiency gains are substantial for large ontologies: selected entities cover under 0.6\% of MeSH and only 6.0\% of MEDIC. For the smaller animal
science ontologies, the percentages are unavoidably higher
(31.9\% for CMO, 35.0\% for VT, 75.2\% for LPT), a smaller subset would yield too few pseudo-labeled pairs for effective fine-tuning; however, the absolute number of processed entities remains modest (1,318, 1,416, and 391, respectively). Furthermore, since aliases are anchored to ontology entities, this supervision is reusable across any corpus linked to the same ontology, whereas per-mention augmentation is not reusable.

To further quantify computational efficiency, Table~\ref{tab:llm_cost} compares Sci-ZSEL with per-mention augmentation in terms of the number of LLM calls and average input tokens per call. Sci-ZSEL requires fewer LLM calls on large ontologies and substantially fewer input tokens per call across all five benchmarks. 

\begin{table}[h]
\setlength{\tabcolsep}{4pt}
\renewcommand{\arraystretch}{1.15}
\centering
\small
\begin{tabular}{lrrrr}
\toprule
& \multicolumn{2}{c}{LLM calls} & \multicolumn{2}{c}{Input tokens} \\
\cmidrule(lr){2-3}\cmidrule(lr){4-5}
Dataset & Ours & Per-ment. & Ours & Per-ment. \\
\midrule
NCBI    & \textbf{805}   & 4,584 & \textbf{89.27} & 302.83 \\
BC5CDR  & \textbf{2,062} & 6,222 & \textbf{77.85} & 300.29 \\
QTL\textsubscript{CMO}& \textbf{1,318} & 1,825 & \textbf{66.07} & 406.73 \\
QTL\textsubscript{VT}  & \textbf{1,416} & 2,134 & \textbf{54.80} & 388.59 \\
QTL\textsubscript{LPT} & \textbf{391}   & 694   & \textbf{50.45} & 395.98 \\
\bottomrule
\end{tabular}
\caption{LLM call counts and average input tokens per call for entity-side selection (Ours, \our) versus a per-mention augmentation baseline. Fewer is better; the lower value in each pair is in bold.}
\label{tab:llm_cost}
\end{table}

\textbf{Alias Lexical Diversity.} Table ~\ref{tab:llm-alias} presents LLM-generated aliases by whether they match a curated synonym (\%Syn), the entity name (\%Name), or neither (constituting a new lexical bridge). On synonym-sparse animal science ontologies (CMO, VT, LPT), both \%Syn and \%Name are notably low. For instance, using the bi-encoder top-1 setting, over 80\% of generated aliases serve as new lexical bridges, successfully providing diverse surface forms that the curated ontologies lack.

\textbf{Pseudo-Pair Accuracy.} 
LLM-based aliases are effective because they create lexical bridges, but they do not work reliably on their own because the raw alias drifts. However, our ontology-aware filter can remove drift. To evaluate the filter's effectiveness on our generated pseudo pairs, we compare against NCBI and BC5CDR, which provide gold labels for the training data. A pseudo-labeled sample is counted as correct whenever its assigned entity matches the gold label. Table~\ref{tab:filter-acc} demonstrates that raw LLM alias generation (\texttt{$P_{\text{Gen}}$w/o filter}) includes noise, yielding a pseudo-label accuracy of ~72\% across both datasets. After removing drifted aliases (\texttt{$P_{\text{Gen}}$}), pseudo pair accuracy rises to over 90\% (90.32\% for NCBI and 91.40\% for BC5CDR). Although pseudo-pair accuracy cannot be directly measured for the animal-science benchmarks because gold training labels are unavailable, their downstream effect is reflected in the EL performance gains.

\subsubsection{Retriever Results}
Table ~\ref{tab:retriever_recall64} presents the retriever's performance, evaluated by overall Recall@64 under three negative-sampling strategies (BASE, RM-PCS, PC-POS), which we compare in Section  ~\ref{sec:retriever_ablation}; Here, we focus on the \our + Synonym results under PC-POS. A key finding is that fine-tuning consistently improves results across all benchmarks, though improvements vary based on synonym sparsity. The magnitude of this improvement is heavily dependent on the dataset's inherent synonym density. On synonym-rich datasets like NCBI and BC5CDR, the non-fine-tuned backbone is already robust, resulting in relatively small fine-tuning gains (+2.68\% and +0.35\%, respectively). On synonym-sparse animal science datasets (QTL\textsubscript{CMO}, QTL\textsubscript{VT}, QTL\textsubscript{LPT}), our ontology-aware fine-tuning delivers substantial improvements of +13.35\% , +15.36\% , and +23.96\%, respectively.

\subsubsection{Reranker Results}
\label{sec:reranker-results}
Table ~\ref{tab:reranker_recall1} presents the reranker results. Fine-tuning improves overall Recall@1 and MRR across every dataset and backbone architecture, where we observe the following major trends: 

When utilizing the BLINK backbone, the combined \our + Synonym approach is clearly the most robust, achieving the highest overall Recall@1 across all five benchmarks. The MRR results closely follow this trend, with \our + Synonym dominating four of the five datasets. The single exception is the QTL\textsubscript{VT} dataset, where the standalone Synonym setting holds a marginal lead of +0.17\% in MRR. For the ReS backbone, the results are more nuanced. \our + Synonym remains the dominant setting for BC5CDR, QTL\textsubscript{VT}, and QTL\textsubscript{LPT}. However, on the NCBI and QTL\textsubscript{CMO} benchmarks, the Synonym setting performs best, outperforming the combined approach in Recall@1 by +7.57\% and +1.05\%, respectively. The MRR results for the ReS backbone mirror this exact trend, with the Synonym baseline maintaining its lead on those same two datasets.

Most notably, \our excels in the highly challenging NO cases across benchmarks, where curated lexical coverage is limited. Fine-tuning drives massive improvements in ZSEL, frequently doubling or tripling the non-fine-tuned backbone performance. Across all five benchmarks, we observe up to a +49.12\%-point absolute gain in the NO category over the non-fine-tuned backbone across BLINK and ReS.

Collectively, these findings confirm that \our effectively addresses a primary challenge of scientific EL. By utilizing LLM-generated aliases, the model successfully captures highly divergent surface forms, providing critical, complementary supervision that manually curated ontology synonyms alone cannot supply.

\subsection{Ablation Studies}
\subsubsection{Retriever Ablations}
\label{sec:retriever_ablation}

We ablate two retriever design choices: the negative-sampling strategy and the pseudo-pair construction strategy. 

For negative sampling, ontology-aware
strategies outperform vanilla in-batch sampling, and PC-POS is the strongest overall, achieving the best Recall@64 on four of five benchmarks (Table~\ref{tab:retriever_recall64}); QTL\textsubscript{VT} is the only exception.

For construction, the union of $\mathcal{P}_1$ and $\mathcal{P}_2$ (\our) with synonyms is the strongest source of pseudo pairs on four of five benchmarks. Together these confirm that integrating ontology structure into both sampling and pseudo-pair construction improves retrieval. Full per-dataset results and discussion are in Appendix~\ref{app:ablation_detail}.

\subsubsection{Reranker Ablations}
We ablate two reranker design choices: the ontology-aware filter and the pseudo-pair construction strategy. 

The filter is necessary: removing it (\our w/o filter) lets drifted aliases push reranker accuracy below the non-fine-tuned baseline on several dataset-backbone combinations (Table~\ref{tab:reranker_recall1}), most severely on the animal science benchmarks.

For construction, the pattern is backbone-dependent, with BLINK, the union (\our) is best or tied-best on four of five benchmarks, whereas with ReS
exact-match pairs alone are more competitive on NCBI and QTL\textsubscript{CMO},
consistent with ReS being more sensitive to alias noise on these datasets. Full
numbers are in Appendix~\ref{app:ablation_detail}.

\section{Acknowledgements} The work is supported in part by NSF-CAREER \#2237831 and USDA-NIFA \#2024-08585.
\section{Conclusion}
In conclusion, this paper offers a cost-aware solution to severe lexical divergence in scientific entity linking. We introduce \our, a zero-shot framework that bounds LLM alias generation to selected entities and employs an ontology-aware filter to discard drifted aliases. To evaluate this, we also release a challenging, high-divergence animal science benchmark (QTL\textsubscript{CMO}, QTL\textsubscript{VT}, QTL\textsubscript{LPT}). Across five datasets, Sci-ZSEL significantly outperforms non-fine-tuned baselines, achieving absolute gains in the lexically divergent domain. We demonstrate that combining \our with curated synonyms yields the most robust reranker configuration, with the gain coming mainly from no-overlap mentions where curated synonyms alone fall short, and our ablation studies confirm the ontology-aware filter is essential to prevent LLM noise from degrading accuracy below baseline levels. Ultimately, this work provides a scalable, reusable blueprint for zero-shot domain adaptation in specialized, low-resource scientific domains.

\clearpage
 \raggedbottom
\section{Limitations}\label{section:limitation}

While \our demonstrates strong performance in scientific entity linking, it is subject to several limitations. First, the framework relies heavily on the domain knowledge of the underlying LLM (Llama 3.2 3B Instruct in our experiments) to generate pseudo pairs. An LLM with weaker biomedical or animal science coverage would produce noisier, more drifted aliases, thereby increasing the burden on the filtering module. Additionally, future adaptations relying on closed-source LLMs could face reproducibility risks if silent updates arbitrarily alter alias distributions. Second, our current framework and evaluation are strictly limited to English. The benchmark datasets, prompt templates, filtering model (BioLORD), and reranking backbones (BLINK, ReS) are all predominantly English-centric. Extending this approach to other languages would require multilingual LLMs and semantic-similarity models specifically tailored to the biomedical domain, which are not currently available off-the-shelf.

Furthermore, the generated pseudo pairs and filtering decisions are intrinsically tied to fixed snapshots of the target ontologies (Appendix Table \ref{tab:onto-versions}). Because ontologies are continuously updated with obsolete entries retired and hierarchical structures, such as the neighbor sets $\mathcal{N}(e)$ and $\mathcal{N}^{+}(e)$, frequently modified—any update to a target ontology requires \our to regenerate the aliases and re-run the filtering process to avoid stale supervision. Finally, our current framework does not handle "NIL" entities, which are mentions lacking a valid referent in the target ontology. Although our new animal science benchmark reveals that a significant portion of mentions cannot be linked to all three ontologies, these NIL cases were excluded from our current evaluation. Developing a robust mechanism to identify and manage unlinkable mentions remains a critical practical challenge that we leave for future work.

\section{Ethical considerations}

\paragraph{Annotation process and compensation.} The animal science benchmark (Section \ref{sec:benchmark}) was developed by a team comprising four graduate-student linking annotators and one domain-expert curator who was responsible for annotation validation and resolving disagreements. Annotation was performed by graduate-student members of the research team and curators of Animal QTLdb.  The annotators were supported by NSF and USDA grants, and were not separately compensated. All source texts were derived entirely from publicly available PubMed articles, ensuring that the dataset contains no personally identifiable information (PII).

\paragraph{LLM usage disclosure.} To generate entity aliases (Section \ref{sec:pseudo-pairs}), we utilized Llama 3.2 3B Instruct, an open-weights large language model. The LLM was prompted strictly using entity definitions sourced from public ontologies; at no point was human-generated or private data processed by the model. All other system components (BLINK, ReS, and BioLORD) are open-source and were deployed in strict adherence to their published usage licenses.

\clearpage

\begingroup
\hypersetup{
    linkcolor=black,
    citecolor=black,
    urlcolor=black
}
\bibliography{custom}
\endgroup

\appendix
\clearpage

\appendix

\section{Appendix}\label{sec:appendix}
\subsection{Entity Generation Prompt}
\label{sec:gen-prompt}

We use a single prompt template across all five datasets, varying
only the domain string and the three in-context examples, both of
which are drawn from the target ontology. The template combines a
system message specifying the domain expertise with a user message
containing the definition to be named:

\begin{quote}
\begin{spacing}{1}
\small\ttfamily
\textbf{[System]} You are a specialist in \{domain\} terminology.
Given an entity's definition, generate a scientifically accurate name
that best represents its meaning. \\[2pt]

Guidelines: \\
1. Carefully interpret the semantic content of the ``definition''. \\
2. Generate a precise name that: \\
\hspace*{1em}- Accurately reflects the definition. \\
\hspace*{1em}- Aligns with standard terminology in \{domain\}. \\[2pt]

Example 1 \\
Definition: \{def$_1$\} \\
Generated name: \{name$_1$\} \\[2pt]

Example 2 \\
Definition: \{def$_2$\} \\
Generated name: \{name$_2$\} \\[2pt]

Example 3 \\
Definition: \{def$_3$\} \\
Generated name: \{name$_3$\} \\[4pt]

\textbf{[User]} Definition : \{definition\} \\
Return all possible names for this definition, separated by commas.
Do not include any additional text.
\end{spacing}
\end{quote}

\noindent The domain string is \emph{biomedical disease} for NCBI
Disease, \emph{biomedical Disease and biomedical Chemical} for
BC5CDR, and \emph{animal science} for QTL\textsubscript{CMO}, QTL\textsubscript{VT}, and QTL\textsubscript{LPT}. The three
in-context examples used for each ontology are listed in
Table~\ref{tab:gen-prompt-examples}.

\FloatBarrier
\subsection{Number of Training Samples}
\label{sec:train-samples}
Table~\ref{tab:smpl_per_setting} reports the number of pseudo-labeled pairs used for retriever and reranker fine-tuning under each setting. The number of samples (SMPL) vary by an order of magnitude across cells, driven by two factors: curated synonym density (Synonym yields 3{,}283 (NCBI), 4{,}676 (BC5CDR), 871 (QTL\textsubscript{CMO}), 1{,}854 (QTL\textsubscript{VT}), and 477 (QTL\textsubscript{LPT}) pairs, tracking the synonym/entity ratios in Table~\ref{tab:dataset-stats} (ontology block)) and filter discard rate (\our retains 74\% (NCBI), 77\% (BC5CDR), 64\% (QTL\textsubscript{CMO}), 29\% (QTL\textsubscript{VT}), and 22\% (QTL\textsubscript{LPT}) of unfiltered pairs, suggesting that the filter removes a larger fraction of generated aliases for the animal science ontologies. \our+Synonym is the largest set across all benchmarks.

\begin{table}[!htbp]
\centering
\setlength{\tabcolsep}{1.5pt}
\renewcommand{\arraystretch}{1.1}
\scriptsize
\begin{tabular}{|l|c|c|c|c|c|}
\hline
\textbf{Setting} & \textbf{NCBI} & \textbf{BC5CDR} & \textbf{QTL\textsubscript{CMO} } & \textbf{QTL\textsubscript{VT} } & \textbf{QTL\textsubscript{LPT} } \\
\hline\hline
% No fine-tune          & --   & --   & --   & --   & --  \\
\our w/o filter   & 1757 & 7447 & 2513 & 1305 & 1592 \\
\our              & 1301 & 5766 & 1614 & 381  & 352 \\
Synonym               & 3283 & 4676 & 871  & 1854 & 477 \\
\our+ Synonym     & 4584 & 6222 & 1825 & 2134 & 694 \\
\hline
\end{tabular}
\caption{Number of training samples (SMPL) per (dataset, setting) used for fine-tuning.}
\label{tab:smpl_per_setting}
\end{table}

\subsection{Bi-encoder Details}
\label{sec:biencoder-appendix}

\subsubsection{Architecture}
\label{sec:biencoder-appendix-arch}
Following \citet{wu2020scalable}, our bi-encoder uses two
independent BERT transformers $T_m$ and $T_e$ to embed the
mention context and the entity into a shared dense space. The
mention input is
\begin{equation}
\tau_m = [\text{CLS}]\, \text{ctx}_l\, [\text{Ms}]\, m\, [\text{Me}]\, \text{ctx}_r\, [\text{SEP}] \:,
\end{equation}
and the entity input is
\begin{equation}
\tau_e = [\text{CLS}]\, \text{title}_e\, [\text{ENT}]\, \text{desc}_e\, [\text{SEP}] \:,
\end{equation}
where $\text{ctx}_l, \text{ctx}_r$ are the left and right contexts
around $m$, and $\text{title}_e, \text{desc}_e$ are the entity title
and description. Each sequence is encoded independently and the
\texttt{[CLS]} output is taken as its dense representation:
\begin{equation}
y_m = \text{red}(T_m(\tau_m)) \:,
\end{equation}
\begin{equation}
y_e = \text{red}(T_e(\tau_e)) \:.
\end{equation}

The mention--entity score is the dot product
\begin{equation}
s(m, e) = y_m \cdot y_e \:.
\label{eq:biencoder-score}
\end{equation}
The bi-encoder is initialized from the BLINK weights of
\citet{wu2020scalable} and fine-tuned on our pseudo-labeled pairs.

\subsubsection{Negative-sampling Formal Definitions}
\label{sec:biencoder-neg-appendix}

\paragraph{Notation.} Let $B = \{m_1, \dots, m_b\}$ be a training
batch of $b$ mentions with gold entities $\{e_1, \dots, e_b\} \subset
\mathcal{E}$. For each batch we construct a candidate column set
$C = G \cup R$, where $G$ is the set of unique golds in the batch
and $R$ is a set of random entities sampled uniformly from
$\mathcal{E}\setminus G$. 

Because the same gold entity may appear for multiple mentions in a batch, removing gold entities from the negative set can reduce the number of available in-batch negatives. We therefore add random entities $R$ sampled from $\mathcal{E}\setminus G$ to keep the number of candidate negatives approximately stable. 
We reuse $\mathcal{N}(e)$ and $\mathcal{N}^{+}(e)$ from
Section~\ref{sec:construction}. For each row $i$ with
gold $e_i$, the positive set $\mathcal{P}_i$ and negative set
$\mathcal{N}_i$ are defined per strategy in
Table~\ref{tab:negsamp-strategies}.

\begin{table}[!htbp]
\centering
\small
\setlength{\tabcolsep}{1pt}
\renewcommand{\arraystretch}{1.1}
\begin{tabular}{|l|c|c|}
\hline
\textbf{Strategy} & \textbf{Positive set $\mathcal{P}_i$} & \textbf{Negative set $\mathcal{N}_i$} \\
\hline\hline
BASE   & $\{e_i\}$ & $(G \setminus \{e_i\}) \cup R$ \\
\hline
RM-PCS & $\{e_i\}$ & $\bigl((G \setminus \{e_i\}) \cup R\bigr) \setminus \mathcal{N}^{+}(e_i)$ \\
\hline
PC-POS & $\{e_i\} \cup \bigl(\mathcal{N}(e_i) \cap G\bigr)$ & $\bigl((G \setminus \mathcal{P}_i) \cup R\bigr) \setminus \mathcal{N}(e_i)$ \\
\hline
\end{tabular}
\caption{Per-row positive set $\mathcal{P}_i$ and negative set $\mathcal{N}_i$ for the three negative-sampling strategies. $\mathcal{N}(e_i) = \text{parents}(e_i) \cup \text{children}(e_i)$; $\mathcal{N}^{+}(e_i) = \mathcal{N}(e_i) \cup \text{siblings}(e_i)$. Siblings are not promoted as positives in PC-POS, preserving sibling separation pressure.}
\label{tab:negsamp-strategies}
\end{table}

\paragraph{Sampling budget.} Naively sampling $|R| = b - |G|$ leaves
$\mathcal{N}_i$ short whenever RM-PCS or PC-POS discards neighbors.
To keep the per-row negative count stable, we sample
$|R| = (b - |G|) + \min(\max_i |\mathcal{N}^{+}(e_i)|, H_{\text{cap}})$
with $H_{\text{cap}} = 30$, then truncate each row's 
$\mathcal{N}_i$ to the target budget.

\subsubsection{Training Objective}
\label{sec:biencoder-obj-appendix}
Following the in-batch cross-entropy objective of
\citet{wu2020scalable}, for each row $i$ the loss is
\begin{equation}
\mathcal{L}_i = -\frac{1}{|\mathcal{P}_i|}\sum_{p \in \mathcal{P}_i}
\log\frac{\exp s(m_i, p)}{Z_i(p)} \:,
\end{equation}
where the partition function is
\begin{equation*}
Z_i(p) = \exp s(m_i, p) + \sum_{c \in \mathcal{N}_i} \exp s(m_i, c)  \:.
\end{equation*}

For BASE and RM-PCS, $|\mathcal{P}_i| = 1$ and the expression
reduces to the standard BLINK in-batch cross-entropy loss; for
PC-POS, the average is taken over multiple positives against the
same negative set. The full batch loss is
\begin{equation}
\mathcal{L} = \frac{1}{b}\sum_{i=1}^{b} \mathcal{L}_i \:.
\end{equation}

\subsection{Cross-encoder Details}
\label{sec:reranker-appendix}

\subsubsection{Architecture}
\label{sec:reranker-arch}
We evaluate two cross-encoder backbones, BLINK
\citep{wu2020scalable} and ReS \citep{xu2023read}. Both share the
same concatenated mention--entity input:
\begin{equation}
\begin{split}
\tau_{m,e} = &[\text{CLS}]\, \text{ctx}_l\, [\text{Ms}]\, m\, [\text{Me}]\, \text{ctx}_r\, [\text{SEP}] \\
&\text{title}_e\, [\text{ENT}]\, \text{desc}_e\, [\text{SEP}] \:,
\end{split}
\end{equation}
where $\text{ctx}_l, \text{ctx}_r$ are the left and right contexts
around $m$, and $\text{title}_e, \text{desc}_e$ are the entity
title and description.

\paragraph{BLINK cross-encoder.} Following \citet{wu2020scalable},
a single BERT transformer $T_{\text{cross}}$ jointly encodes
$\tau_{m,e}$, and the score is a linear projection of the reduced
[CLS] embedding:
\begin{equation}
y_{m,e} = \text{red}(T_{\text{cross}}(\tau_{m,e}))\:,
\end{equation}
\begin{equation}
s_{\text{BLINK}}(m, e) = \mathbf{w}^\top y_{m,e} \:,
\end{equation}
where $\mathbf{w}$ is a learned scoring vector. The BLINK reranker
is initialized from the cross-encoder weights of
\citet{wu2020scalable} pretrained on Wikipedia entity linking.

\paragraph{ReS cross-encoder.} The Read-and-Select framework
\citep{xu2023read} factors scoring into two stages. In the
\emph{reading} stage, each candidate $e \in \mathcal{C}(m)$ is
encoded with the mention by a cross-encoder $T_{\text{read}}$ to
produce a candidate-conditioned mention representation $h_m^e$.
In the \emph{selecting} stage, the $|\mathcal{C}(m)|$
representations are passed through a candidate-level transformer
that performs cross-candidate attention, so the score for $e$
depends on the other candidates in $\mathcal{C}(m)$ rather than on
$e$ alone:
\begin{equation}
s_{\text{ReS}}(m, e) = f_{\text{sel}}\!\bigl(\{h_m^{e'}\}_{e' \in \mathcal{C}(m)},\, e\bigr) \:,
\end{equation}
where $f_{\text{sel}}$ is the selecting head; see
\citet{xu2023read} for the full architecture. The ReS reranker is
initialized from the released ReS weights.

\subsubsection{Candidate Set Construction}
\label{sec:candidate-set}
For each pseudo pair $\langle m, e \rangle$, let $\{c_1, \ldots, c_K\}$ be the top-$K$ candidates from the BLINK bi-encoder, ranked by retriever score. The reranker candidate set $\mathcal{C}(m)$ equals $\{c_1, \ldots, c_K\}$ when $e \in \{c_1, \ldots, c_K\}$, and otherwise replaces the lowest-ranked candidate with $e$, giving $\{c_1, \ldots, c_{K-1}, e\}$. This guarantees $e \in \mathcal{C}(m)$ and $|\mathcal{C}(m)| = K$. 

We set K=64. After ensuring that the pseudo-labeled entity $e$ is included in the candidate set, $e$ is treated as the single positive and the remaining 63 candidates as negatives. For reranker fine-tuning, we randomly sample 20 negative examples from these 63 negatives and train the reranker using the positive entity together with the 20 sampled negatives. The reranker is then trained on the reduced candidate set whereas the full candidate set is retained at inference, where the reranker scores all $K$ retrieved candidates.

\subsubsection{Training Objective}
\label{sec:reranker-obj}
For each pseudo pair $\langle m, e \rangle$ with candidate set
$\mathcal{C}(m)$, the reranker is trained with cross-entropy,
treating $e$ as the positive and the remaining candidates as
negatives:
\begin{equation}
\mathcal{L} = -\log\frac{\exp s(m, e)}{\sum_{c \in \mathcal{C}(m)} \exp s(m, c)} \:,
\end{equation}
where $s(m, c)$ is the reranker score ($s_{\text{BLINK}}$ or
$s_{\text{ReS}}$). The same objective is used for both backbones;
only the scoring function differs.

\subsection{Training Hyperparameters and Hardware}
\label{app:hyperparameters}

\paragraph{Hardware.} LLM alias generation uses
Llama~3.2~3B Instruct on a single NVIDIA A100. Retriever
and reranker fine-tuning use four A100s.

\paragraph{Bi-encoder retriever.} The BLINK retriever uses
\texttt{bert-large-uncased} with a 128-token mention context
and 128-token candidate. It is trained at a learning rate of
$2\times 10^{-5}$ with batch size 512, falling back to 128
when fewer than 512 training pairs are available.

\paragraph{Cross-encoder rerankers.} The BLINK reranker uses
\texttt{bert-large-uncased} with a 64-token context and
128-token candidate, trained at learning rate
$2\times 10^{-5}$ with batch size 32. The ReS reranker uses
\texttt{roberta-base} with a 256-token context and 256-token
entity description, trained at learning rate $1\times 10^{-4}$
with batch size 8. Dropout is fixed at 0.2 throughout.

\paragraph{Epochs and reporting protocol.} Because the zero-shot setting provides no validation set, best-epoch
selection is not possible; we fix the number of epochs in
advance and report results from the last epoch. The retriever
runs for 4 epochs on BC5CDR (the largest dataset) and 1 epoch
on the remaining four datasets. The reranker runs for 3 epochs
on all benchmarks. Empirical loss curves confirm that the chosen epoch counts place the final epoch at or near the training optimum for nearly all
benchmark--backbone--seed combinations,\footnote{The only exception is the ReS reranker on BC5CDR under $\mathcal{P}_1 +$ Synonym. One of the three seeds becomes unstable after epoch~1: the training loss increases, while test accuracy drops from about $75\%$ to $10\%$. Following our fixed-epoch protocol, we report this run without modification. This unstable seed leads to the high variance ($50.98 \pm 41.58$) reported for this setting in Table~\ref{tab:construction_ablation}, while the other two seeds remain stable.} validating our last-epoch
reporting protocol.

\subsection{Ontology/Knowledge Base Versions}
\label{sec:onto-versions}

Table~\ref{tab:onto-versions} reports the specific releases of each
knowledge base used in our experiments. Versions were fixed at the
start of experimentation and reused across all settings.

\begin{table}[h]
\centering
\small
\begin{tabular}{lll}
\toprule
KB / Ontology & Used by & Release \\
\midrule
MeSH 2025     & BC5CDR & 2025-01-01 \\
MEDIC         & NCBI Disease & 2025-02-28 \\
CMO           & QTL\textsubscript{CMO}      & 2026-02-28 \\
VT            & QTL\textsubscript{VT}      & 2026-03-03 \\
LPT           & QTL\textsubscript{LPT}      & 2025-08-14 \\
\bottomrule
\end{tabular}
\caption{Release versions of the knowledge bases and ontologies used
in this work.}
\label{tab:onto-versions}
\end{table}

\subsection{Detailed Ablation Analysis}
\label{app:ablation_detail}

\paragraph{Retriever: negative sampling.}
We compare BASE, RM-PCS, and PC-POS under the Sci-ZSEL~+~Synonym framework
(Table~\ref{tab:retriever_recall64}). PC-POS achieves the highest Recall@64 on NCBI, BC5CDR, QTL\textsubscript{CMO}, and QTL\textsubscript{LPT}. The sole exception is QTL\textsubscript{VT}, where RM-PCS peaks at 88.69\%; for VT's ontological
structure, removing sibling entities from the negatives (RM-PCS) yields a
stronger signal than adding neighbor entities as positives (PC-POS). The MRR
results (Tables~\ref{tab:recall64_base}, ~\ref{tab:recall64_rmpcs}, \ref{tab:recall64_pcpos}) follow a broadly similar trend: PC-POS achieves the highest MRR on NCBI, BC5CDR, and QTL\textsubscript{LPT}, RM-PCS on QTL\textsubscript{VT}, and BASE on QTL\textsubscript{CMO}. These results confirm that integrating ontology structure into sampling is superior to vanilla construction.

\paragraph{Retriever: construction strategy.}
We ablate the pseudo-pair construction strategy in Table~\ref{tab:construction_ablation}. \our~+~Synonym is the best strategy on
NCBI, BC5CDR, QTL\textsubscript{CMO}, and QTL\textsubscript{LPT}, showing that the
union of $\mathcal{P}_1$ (Eq.~\ref{eq:p1}) and $\mathcal{P}_2$ (Eq.~\ref{eq:p2}) with synonyms 
is the strongest source of pseudo pairs. QTL\textsubscript{VT} is the only
exception, where $\mathcal{P}_2$~+~Synonym leads by a small margin
(89.18\% vs.\ 87.99\%). Both construction strategies contribute to retrieval, and
combining them is generally preferable.

\paragraph{Reranker: ontology-aware filter.}
We verify the necessity of the filter by omitting it (\our w/o filter).
Removing the filter often harms performance, causing reranker accuracy
(Table~\ref{tab:reranker_recall1}) to drop below the non-fine-tuned baseline across
several dataset--backbone combinations. The largest drop in overall Recall@1 is
7.50\% on QTL\textsubscript{CMO} with ReS. This occurs because drifted aliases
train the model to link mentions to a neighboring entity rather than the true
referent; as a result, performance drops even on the easiest HO cases (the
largest HO Recall@1 drop is 37.81\%). These effects are most severe on the animal
science datasets. We conclude that the ontology-aware filter is necessary to
leverage LLM-based aliases, especially in low-resource settings.

\paragraph{Reranker: construction strategy.}
We ablate construction under both backbones in
Table~\ref{tab:construction_ablation}. With BLINK, \our~+~Synonym is best or tied-best on four of five datasets, again favoring the union of $\mathcal{P}_1$ and $\mathcal{P}_2$. With ReS, exact-match alone ($\mathcal{P}_1$~+~Synonym) is more competitive on NCBI and QTL\textsubscript{CMO}, and on these two datasets ReS also performs best using ontology synonyms alone (Table~\ref{tab:reranker_recall1}). This suggests ReS is more sensitive to training noise on these datasets, so the added LLM aliases help less there.

\begin{table*}[t]
\centering
\setlength{\tabcolsep}{6pt}
\renewcommand{\arraystretch}{1.1}
\scriptsize
\begin{tabular}{|c|l|c|c|c|c|c|}
\hline
\textbf{Dataset} & \textbf{Setting} & \textbf{Overall} & \textbf{MRR} & \textbf{HO} & \textbf{LO} & \textbf{NO} \\
\hline\hline
\multirow{5}{*}{\textbf{NCBI}} & No fine-tune & 88.12$_{\pm 0.00}$ & 61.94$_{\pm 0.00}$ & \textbf{99.68$_{\pm 0.00}$} & 92.82$_{\pm 0.00}$ & 62.73$_{\pm 0.00}$ \\
 & \our w/o filter & 87.74$_{\pm 0.22}$ & 62.48$_{\pm 1.82}$ & 99.46$_{\pm 0.19}$ & 91.67$_{\pm 0.23}$ & 63.64$_{\pm 1.58}$ \\
 & \our & 87.40$_{\pm 0.47}$ & 58.11$_{\pm 2.28}$ & \textbf{99.68$_{\pm 0.00}$} & 90.97$_{\pm 0.61}$ & 63.18$_{\pm 1.21}$ \\
 & Synonym & 89.20$_{\pm 0.24}$ & \textbf{64.53$_{\pm 2.28}$} & 99.46$_{\pm 0.38}$ & 92.05$_{\pm 0.35}$ & 69.24$_{\pm 0.95}$ \\
 & \our+ Synonym & \textbf{89.72$_{\pm 0.97}$} & 64.35$_{\pm 2.10}$ & \textbf{99.68$_{\pm 0.00}$} & \textbf{92.90$_{\pm 1.33}$} & \textbf{69.54$_{\pm 1.64}$} \\
\hline
\multirow{5}{*}{\textbf{BC5CDR}} & No fine-tune & \textbf{88.79$_{\pm 0.00}$} & 73.91$_{\pm 0.00}$ & 99.46$_{\pm 0.00}$ & \textbf{96.09$_{\pm 0.00}$} & \textbf{57.24$_{\pm 0.00}$} \\
 & \our w/o filter & 88.41$_{\pm 0.74}$ & \textbf{74.09$_{\pm 1.22}$} & \textbf{99.49$_{\pm 0.07}$} & 95.52$_{\pm 1.05}$ & 55.88$_{\pm 2.40}$ \\
 & \our & 85.27$_{\pm 1.75}$ & 70.75$_{\pm 2.08}$ & 99.23$_{\pm 0.23}$ & 89.20$_{\pm 3.75}$ & 47.17$_{\pm 4.53}$ \\
 & Synonym & 87.22$_{\pm 0.16}$ & 72.03$_{\pm 0.26}$ & 99.25$_{\pm 0.10}$ & 91.64$_{\pm 1.08}$ & 53.78$_{\pm 0.45}$ \\
 & \our+ Synonym & 87.96$_{\pm 0.48}$ & 73.20$_{\pm 0.33}$ & 99.40$_{\pm 0.04}$ & 93.55$_{\pm 1.01}$ & 55.39$_{\pm 1.60}$ \\
\hline
\multirow{5}{*}{\textbf{QTL\textsubscript{CMO} }} & No fine-tune & 75.74$_{\pm 0.00}$ & 43.20$_{\pm 0.00}$ & 99.27$_{\pm 0.00}$ & 87.43$_{\pm 0.00}$ & 55.04$_{\pm 0.00}$ \\
 & \our w/o filter & \textbf{89.96$_{\pm 0.23}$} & 54.09$_{\pm 2.73}$ & \textbf{99.51$_{\pm 0.00}$} & 97.48$_{\pm 0.08}$ & \textbf{79.24$_{\pm 0.47}$} \\
 & \our & 89.22$_{\pm 0.86}$ & 56.06$_{\pm 0.55}$ & \textbf{99.51$_{\pm 0.00}$} & 97.97$_{\pm 0.24}$ & 77.12$_{\pm 2.16}$ \\
 & Synonym & 85.71$_{\pm 0.40}$ & 53.66$_{\pm 0.89}$ & 99.35$_{\pm 0.14}$ & 97.12$_{\pm 0.39}$ & 69.84$_{\pm 0.54}$ \\
 & \our+ Synonym & 88.22$_{\pm 0.89}$ & \textbf{58.17$_{\pm 1.01}$} & \textbf{99.51$_{\pm 0.00}$} & \textbf{98.29$_{\pm 0.20}$} & 74.56$_{\pm 1.93}$ \\
\hline
\multirow{5}{*}{\textbf{QTL\textsubscript{VT} }} & No fine-tune & 72.63$_{\pm 0.00}$ & 34.67$_{\pm 0.00}$ & \textbf{99.33$_{\pm 0.00}$} & 86.96$_{\pm 0.00}$ & 59.96$_{\pm 0.00}$ \\
 & \our w/o filter & 84.70$_{\pm 0.96}$ & 43.95$_{\pm 1.20}$ & \textbf{99.33$_{\pm 0.00}$} & 96.64$_{\pm 0.44}$ & 75.31$_{\pm 1.53}$ \\
 & \our & 84.72$_{\pm 1.69}$ & 45.20$_{\pm 0.90}$ & \textbf{99.33$_{\pm 0.00}$} & \textbf{96.75$_{\pm 0.40}$} & 75.28$_{\pm 2.73}$ \\
 & Synonym & \textbf{86.99$_{\pm 1.74}$} & \textbf{47.83$_{\pm 4.78}$} & \textbf{99.33$_{\pm 0.00}$} & 93.22$_{\pm 2.30}$ & \textbf{81.36$_{\pm 1.82}$} \\
 & \our+ Synonym & 85.58$_{\pm 4.10}$ & 45.50$_{\pm 15.06}$ & 98.43$_{\pm 1.55}$ & 91.88$_{\pm 4.37}$ & 79.84$_{\pm 4.55}$ \\
\hline
\multirow{5}{*}{\textbf{QTL\textsubscript{LPT} }} & No fine-tune & 68.70$_{\pm 0.00}$ & 51.17$_{\pm 0.00}$ & \textbf{100.00$_{\pm 0.00}$} & 90.86$_{\pm 0.00}$ & 35.38$_{\pm 0.00}$ \\
 & \our w/o filter & 70.08$_{\pm 0.28}$ & 47.50$_{\pm 0.96}$ & \textbf{100.00$_{\pm 0.00}$} & 93.33$_{\pm 0.33}$ & 37.13$_{\pm 0.47}$ \\
 & \our & 75.07$_{\pm 1.54}$ & 54.76$_{\pm 0.14}$ & \textbf{100.00$_{\pm 0.00}$} & \textbf{96.19$_{\pm 0.33}$} & 46.67$_{\pm 3.46}$ \\
 & Synonym & \textbf{92.43$_{\pm 0.68}$} & \textbf{75.56$_{\pm 0.87}$} & \textbf{100.00$_{\pm 0.00}$} & 91.05$_{\pm 2.01}$ & \textbf{88.00$_{\pm 0.62}$} \\
 & \our+ Synonym & 92.29$_{\pm 0.44}$ & 57.38$_{\pm 1.37}$ & \textbf{100.00$_{\pm 0.00}$} & 92.19$_{\pm 0.66}$ & 87.08$_{\pm 0.93}$ \\
\hline
\end{tabular}
\caption{Retriever Recall@64 under the \textbf{BASE} negative-sampling strategy. \textbf{Bold} represents the best results when comparing row-wise.}
\label{tab:recall64_base}
\end{table*}

\begin{table*}[t]
\centering
\setlength{\tabcolsep}{6pt}
\renewcommand{\arraystretch}{1.1}
\scriptsize
\begin{tabular}{|c|l|c|c|c|c|c|}
\hline
\textbf{Dataset} & \textbf{Setting} & \textbf{Overall} & \textbf{MRR} & \textbf{HO} & \textbf{LO} & \textbf{NO} \\
\hline\hline
\multirow{5}{*}{\textbf{NCBI}} & No fine-tune & 88.12$_{\pm 0.00}$ & 61.94$_{\pm 0.00}$ & \textbf{99.68$_{\pm 0.00}$} & \textbf{92.82$_{\pm 0.00}$} & 62.73$_{\pm 0.00}$ \\
 & \our w/o filter & 87.01$_{\pm 1.85}$ & 63.75$_{\pm 1.48}$ & 98.70$_{\pm 1.42}$ & 90.67$_{\pm 1.74}$ & 63.48$_{\pm 2.96}$ \\
 & \our & 80.70$_{\pm 11.97}$ & 54.37$_{\pm 10.62}$ & 92.21$_{\pm 12.94}$ & 83.26$_{\pm 13.97}$ & 59.55$_{\pm 6.70}$ \\
 & Synonym & \textbf{89.73$_{\pm 0.30}$} & 63.99$_{\pm 1.16}$ & \textbf{99.68$_{\pm 0.00}$} & 92.36$_{\pm 1.16}$ & \textbf{70.60$_{\pm 1.31}$} \\
 & \our+ Synonym & 89.41$_{\pm 0.78}$ & \textbf{64.41$_{\pm 1.94}$} & \textbf{99.68$_{\pm 0.00}$} & 91.66$_{\pm 0.80}$ & \textbf{70.60$_{\pm 2.15}$} \\
\hline
\multirow{5}{*}{\textbf{BC5CDR}} & No fine-tune & \textbf{88.79$_{\pm 0.00}$} & \textbf{73.91$_{\pm 0.00}$} & \textbf{99.46$_{\pm 0.00}$} & \textbf{96.09$_{\pm 0.00}$} & \textbf{57.24$_{\pm 0.00}$} \\
 & \our w/o filter & 88.21$_{\pm 2.01}$ & 73.51$_{\pm 2.57}$ & 99.42$_{\pm 0.23}$ & 95.39$_{\pm 3.00}$ & 55.30$_{\pm 6.11}$ \\
 & \our & 85.41$_{\pm 0.27}$ & 70.29$_{\pm 0.66}$ & 99.29$_{\pm 0.15}$ & 89.20$_{\pm 1.98}$ & 47.63$_{\pm 2.14}$ \\
 & Synonym & 87.13$_{\pm 0.59}$ & 72.24$_{\pm 0.58}$ & 99.34$_{\pm 0.19}$ & 91.87$_{\pm 0.86}$ & 53.08$_{\pm 2.85}$ \\
 & \our+ Synonym & 86.61$_{\pm 2.16}$ & 68.05$_{\pm 7.57}$ & 99.00$_{\pm 0.56}$ & 90.18$_{\pm 5.00}$ & 52.75$_{\pm 4.63}$ \\
\hline
\multirow{5}{*}{\textbf{QTL\textsubscript{CMO} }} & No fine-tune & 75.74$_{\pm 0.00}$ & 43.20$_{\pm 0.00}$ & 99.27$_{\pm 0.00}$ & 87.43$_{\pm 0.00}$ & 55.04$_{\pm 0.00}$ \\
 & \our w/o filter & 88.02$_{\pm 1.24}$ & 54.37$_{\pm 1.05}$ & \textbf{99.51$_{\pm 0.00}$} & 95.68$_{\pm 2.39}$ & 76.29$_{\pm 0.86}$ \\
 & \our & 88.06$_{\pm 0.25}$ & 56.00$_{\pm 0.95}$ & \textbf{99.51$_{\pm 0.00}$} & 97.97$_{\pm 0.24}$ & 74.44$_{\pm 0.51}$ \\
 & Synonym & 85.47$_{\pm 0.49}$ & 53.47$_{\pm 0.35}$ & 99.27$_{\pm 0.00}$ & 97.07$_{\pm 0.16}$ & 69.35$_{\pm 1.05}$ \\
 & \our+ Synonym & \textbf{89.03$_{\pm 1.68}$} & \textbf{57.17$_{\pm 0.99}$} & \textbf{99.51$_{\pm 0.00}$} & \textbf{98.33$_{\pm 0.34}$} & \textbf{76.37$_{\pm 3.58}$} \\
\hline
\multirow{5}{*}{\textbf{QTL\textsubscript{VT} }} & No fine-tune & 72.63$_{\pm 0.00}$ & 34.67$_{\pm 0.00}$ & \textbf{99.33$_{\pm 0.00}$} & 86.96$_{\pm 0.00}$ & 59.96$_{\pm 0.00}$ \\
 & \our w/o filter & 84.52$_{\pm 1.08}$ & 44.05$_{\pm 0.51}$ & \textbf{99.33$_{\pm 0.00}$} & 96.41$_{\pm 0.36}$ & 75.14$_{\pm 1.79}$ \\
 & \our & 85.51$_{\pm 0.30}$ & 43.49$_{\pm 0.72}$ & \textbf{99.33$_{\pm 0.00}$} & \textbf{96.64$_{\pm 0.44}$} & 76.73$_{\pm 0.26}$ \\
 & Synonym & 86.79$_{\pm 2.67}$ & 46.21$_{\pm 3.65}$ & \textbf{99.33$_{\pm 0.00}$} & 93.04$_{\pm 3.31}$ & 81.12$_{\pm 2.81}$ \\
 & \our+ Synonym & \textbf{88.69$_{\pm 2.22}$} & \textbf{49.50$_{\pm 4.52}$} & \textbf{99.33$_{\pm 0.00}$} & 94.32$_{\pm 3.26}$ & \textbf{83.68$_{\pm 1.95}$} \\
\hline
\multirow{5}{*}{\textbf{QTL\textsubscript{LPT} }} & No fine-tune & 68.70$_{\pm 0.00}$ & 51.17$_{\pm 0.00}$ & \textbf{100.00$_{\pm 0.00}$} & 90.86$_{\pm 0.00}$ & 35.38$_{\pm 0.00}$ \\
 & \our w/o filter & 70.82$_{\pm 0.97}$ & 48.46$_{\pm 1.21}$ & \textbf{100.00$_{\pm 0.00}$} & 93.71$_{\pm 0.58}$ & 38.57$_{\pm 1.85}$ \\
 & \our & 76.87$_{\pm 0.73}$ & 54.03$_{\pm 1.07}$ & \textbf{100.00$_{\pm 0.00}$} & \textbf{95.24$_{\pm 1.19}$} & 51.18$_{\pm 2.00}$ \\
 & Synonym & 91.83$_{\pm 0.42}$ & \textbf{75.45$_{\pm 1.19}$} & \textbf{100.00$_{\pm 0.00}$} & 90.10$_{\pm 1.84}$ & 87.18$_{\pm 0.35}$ \\
 & \our+ Synonym & \textbf{92.48$_{\pm 0.35}$} & 57.37$_{\pm 0.59}$ & \textbf{100.00$_{\pm 0.00}$} & 92.57$_{\pm 1.14}$ & \textbf{87.28$_{\pm 0.17}$} \\
\hline
\end{tabular}
\caption{Retriever Recall@64 under the \textbf{RM-PCS} negative-sampling strategy. \textbf{Bold} represents the best results when comparing row-wise.}
\label{tab:recall64_rmpcs}
\end{table*}

\begin{table*}[t]
\centering
\setlength{\tabcolsep}{6pt}
\renewcommand{\arraystretch}{1.1}
\scriptsize
\begin{tabular}{|c|l|c|c|c|c|c|}
\hline
\textbf{Dataset} & \textbf{Setting} & \textbf{Overall} & \textbf{MRR} & \textbf{HO} & \textbf{LO} & \textbf{NO} \\
\hline\hline
\multirow{5}{*}{\textbf{NCBI}} & No fine-tune & 88.12$_{\pm 0.00}$ & 61.94$_{\pm 0.00}$ & \textbf{99.68$_{\pm 0.00}$} & \textbf{92.82$_{\pm 0.00}$} & 62.73$_{\pm 0.00}$ \\
 & \our w/o filter & 88.75$_{\pm 0.36}$ & 63.78$_{\pm 1.61}$ & 99.46$_{\pm 0.19}$ & 92.21$_{\pm 0.53}$ & 66.97$_{\pm 2.89}$ \\
 & \our & 87.81$_{\pm 0.63}$ & 61.12$_{\pm 1.48}$ & \textbf{99.68$_{\pm 0.00}$} & 91.82$_{\pm 0.93}$ & 63.33$_{\pm 0.95}$ \\
 & Synonym & 90.21$_{\pm 0.36}$ & 65.20$_{\pm 0.97}$ & 99.35$_{\pm 0.33}$ & 91.98$_{\pm 0.48}$ & 73.94$_{\pm 0.26}$ \\
 & \our+ Synonym & \textbf{90.80$_{\pm 0.87}$} & \textbf{66.63$_{\pm 1.46}$} & 99.46$_{\pm 0.38}$ & 91.98$_{\pm 0.48}$ & \textbf{76.37$_{\pm 2.36}$} \\
\hline
\multirow{5}{*}{\textbf{BC5CDR}} & No fine-tune & 88.79$_{\pm 0.00}$ & 73.91$_{\pm 0.00}$ & 99.46$_{\pm 0.00}$ & 96.09$_{\pm 0.00}$ & 57.24$_{\pm 0.00}$ \\
 & \our w/o filter & 88.93$_{\pm 1.03}$ & 73.76$_{\pm 1.74}$ & \textbf{99.53$_{\pm 0.04}$} & \textbf{96.19$_{\pm 1.62}$} & 57.54$_{\pm 3.34}$ \\
 & \our & 87.80$_{\pm 0.74}$ & 72.63$_{\pm 0.97}$ & 99.42$_{\pm 0.10}$ & 94.60$_{\pm 1.50}$ & 54.09$_{\pm 2.09}$ \\
 & Synonym & 88.57$_{\pm 0.50}$ & 73.50$_{\pm 0.51}$ & 99.48$_{\pm 0.05}$ & 95.50$_{\pm 0.62}$ & 56.58$_{\pm 2.24}$ \\
 & \our+ Synonym & \textbf{89.14$_{\pm 0.65}$} & \textbf{74.05$_{\pm 0.46}$} & 99.29$_{\pm 0.20}$ & 95.60$_{\pm 1.58}$ & \textbf{59.38$_{\pm 1.69}$} \\
\hline
\multirow{5}{*}{\textbf{QTL\textsubscript{CMO} }} & No fine-tune & 75.74$_{\pm 0.00}$ & 43.20$_{\pm 0.00}$ & 99.27$_{\pm 0.00}$ & 87.43$_{\pm 0.00}$ & 55.04$_{\pm 0.00}$ \\
 & \our w/o filter & \textbf{89.24$_{\pm 1.61}$} & 54.53$_{\pm 1.76}$ & \textbf{99.51$_{\pm 0.00}$} & 96.89$_{\pm 1.18}$ & \textbf{78.07$_{\pm 2.77}$} \\
 & \our & 88.34$_{\pm 1.37}$ & 54.79$_{\pm 1.08}$ & \textbf{99.51$_{\pm 0.00}$} & 97.79$_{\pm 0.16}$ & 75.23$_{\pm 3.14}$ \\
 & Synonym & 85.45$_{\pm 0.91}$ & 53.31$_{\pm 0.96}$ & 99.43$_{\pm 0.14}$ & 97.16$_{\pm 0.41}$ & 69.16$_{\pm 1.71}$ \\
 & \our+ Synonym & 89.09$_{\pm 1.57}$ & \textbf{56.33$_{\pm 0.35}$} & \textbf{99.51$_{\pm 0.00}$} & \textbf{98.33$_{\pm 0.20}$} & 76.52$_{\pm 3.46}$ \\
\hline
\multirow{5}{*}{\textbf{QTL\textsubscript{VT} }} & No fine-tune & 72.63$_{\pm 0.00}$ & 34.67$_{\pm 0.00}$ & \textbf{99.33$_{\pm 0.00}$} & 86.96$_{\pm 0.00}$ & 59.96$_{\pm 0.00}$ \\
 & \our w/o filter & 83.69$_{\pm 0.94}$ & 43.61$_{\pm 0.66}$ & \textbf{99.33$_{\pm 0.00}$} & 96.23$_{\pm 0.36}$ & 73.79$_{\pm 1.51}$ \\
 & \our & 85.31$_{\pm 1.03}$ & 45.32$_{\pm 0.93}$ & \textbf{99.33$_{\pm 0.00}$} & \textbf{96.70$_{\pm 0.30}$} & 76.35$_{\pm 1.67}$ \\
 & Synonym & 87.91$_{\pm 2.22}$ & 46.94$_{\pm 2.40}$ & \textbf{99.33$_{\pm 0.00}$} & 93.33$_{\pm 1.86}$ & 82.92$_{\pm 2.94}$ \\
 & \our+ Synonym & \textbf{87.99$_{\pm 1.47}$} & \textbf{47.63$_{\pm 2.73}$} & \textbf{99.33$_{\pm 0.00}$} & 92.69$_{\pm 3.08}$ & \textbf{83.44$_{\pm 1.24}$} \\
\hline
\multirow{5}{*}{\textbf{QTL\textsubscript{LPT} }} & No fine-tune & 68.70$_{\pm 0.00}$ & 51.17$_{\pm 0.00}$ & \textbf{100.00$_{\pm 0.00}$} & 90.86$_{\pm 0.00}$ & 35.38$_{\pm 0.00}$ \\
 & \our w/o filter & 71.01$_{\pm 0.92}$ & 50.47$_{\pm 1.78}$ & \textbf{100.00$_{\pm 0.00}$} & 94.86$_{\pm 1.51}$ & 38.36$_{\pm 1.24}$ \\
 & \our & 76.18$_{\pm 0.37}$ & 54.01$_{\pm 0.29}$ & \textbf{100.00$_{\pm 0.00}$} & \textbf{95.62$_{\pm 1.19}$} & 49.44$_{\pm 1.24}$ \\
 & Synonym & 91.78$_{\pm 1.24}$ & \textbf{75.72$_{\pm 0.62}$} & \textbf{100.00$_{\pm 0.00}$} & 90.28$_{\pm 2.49}$ & 86.97$_{\pm 1.43}$ \\
 & \our+ Synonym & \textbf{92.66$_{\pm 0.69}$} & 57.61$_{\pm 0.65}$ & \textbf{100.00$_{\pm 0.00}$} & 93.14$_{\pm 1.15}$ & \textbf{87.38$_{\pm 0.93}$} \\
\hline
\end{tabular}
\caption{Retriever Recall@64 under the \textbf{PC-POS} negative-sampling strategy. \textbf{Bold} represents the best results when comparing row-wise.}
\label{tab:recall64_pcpos}
\end{table*}

% ===== Merged construction-strategy ablation (retriever + reranker) =====
\begin{table*}[t]
\centering
\setlength{\tabcolsep}{6pt}
\renewcommand{\arraystretch}{1.1}
\scriptsize
\begin{tabular}{|c|l|c|c|c|c|c|}
\hline
\multirow{2}{*}{\textbf{Dataset}} & \multirow{2}{*}{\textbf{Setting}} & \textbf{Retriever} & \multicolumn{2}{c|}{\textbf{BLINK reranker}} & \multicolumn{2}{c|}{\textbf{ReS reranker}} \\
\cline{3-7}
 & & \textbf{Recall@64} & \textbf{Recall@1} & \textbf{MRR} & \textbf{Recall@1} & \textbf{MRR} \\
\hline
\hline
\multirow{3}{*}{\textbf{NCBI}} & $\mathcal{P}_1$ + Synonym & 90.56$_{\pm 0.16}$ & 77.15$_{\pm 0.73}$ & 83.88$_{\pm 0.36}$ & \textbf{78.26$_{\pm 0.91}$} & \textbf{84.46$_{\pm 0.79}$} \\
 & $\mathcal{P}_2$ + Synonym & 90.42$_{\pm 0.68}$ & 74.93$_{\pm 4.49}$ & 82.95$_{\pm 2.22}$ & 70.69$_{\pm 0.57}$ & 80.06$_{\pm 1.07}$ \\
 & \our + Synonym & \textbf{90.80$_{\pm 0.87}$} & \textbf{77.43$_{\pm 0.67}$} & \textbf{84.35$_{\pm 0.74}$} & 70.52$_{\pm 1.02}$ & 79.11$_{\pm 1.58}$ \\
\hline
\multirow{3}{*}{\textbf{BC5CDR}} & $\mathcal{P}_1$ + Synonym & 88.47$_{\pm 0.35}$ & 79.35$_{\pm 0.38}$ & 84.98$_{\pm 0.42}$ & 50.98$_{\pm 41.58}$ & 56.75$_{\pm 40.31}$ \\
 & $\mathcal{P}_2$ + Synonym & 88.71$_{\pm 0.83}$ & 81.19$_{\pm 0.59}$ & \textbf{86.33$_{\pm 0.65}$} & 77.37$_{\pm 0.67}$ & 82.91$_{\pm 0.62}$ \\
 & \our + Synonym & \textbf{89.14$_{\pm 0.65}$} & \textbf{81.20$_{\pm 0.58}$} & 86.20$_{\pm 0.48}$ & \textbf{77.90$_{\pm 0.12}$} & \textbf{83.21$_{\pm 0.26}$} \\
\hline
\multirow{3}{*}{\textbf{QTL\textsubscript{CMO} }} & $\mathcal{P}_1$ + Synonym & 87.45$_{\pm 0.50}$ & 57.63$_{\pm 2.02}$ & 66.97$_{\pm 1.85}$ & \textbf{58.60$_{\pm 1.94}$} & \textbf{68.09$_{\pm 1.63}$} \\
 & $\mathcal{P}_2$ + Synonym & 87.01$_{\pm 0.60}$ & \textbf{59.28$_{\pm 0.36}$} & \textbf{68.27$_{\pm 0.44}$} & 57.67$_{\pm 0.92}$ & 67.82$_{\pm 0.81}$ \\
 & \our + Synonym & \textbf{89.09$_{\pm 1.57}$} & 58.88$_{\pm 3.30}$ & 67.72$_{\pm 2.92}$ & 57.43$_{\pm 0.23}$ & 66.95$_{\pm 0.12}$ \\
\hline
\multirow{3}{*}{\textbf{QTL\textsubscript{VT} }} & $\mathcal{P}_1$ + Synonym & 88.47$_{\pm 0.80}$ & 69.43$_{\pm 3.89}$ & 78.42$_{\pm 2.75}$ & 61.30$_{\pm 4.36}$ & 71.92$_{\pm 3.16}$ \\
 & $\mathcal{P}_2$ + Synonym & \textbf{89.18$_{\pm 0.63}$} & 70.58$_{\pm 1.49}$ & \textbf{79.75$_{\pm 0.81}$} & 65.30$_{\pm 3.67}$ & 75.32$_{\pm 3.15}$ \\
 & \our + Synonym & 87.99$_{\pm 1.47}$ & \textbf{71.25$_{\pm 2.58}$} & 79.49$_{\pm 1.65}$ & \textbf{65.58$_{\pm 2.62}$} & \textbf{75.76$_{\pm 2.03}$} \\
\hline
\multirow{3}{*}{\textbf{QTL\textsubscript{LPT} }} & $\mathcal{P}_1$ + Synonym & 92.56$_{\pm 0.44}$ & 79.36$_{\pm 1.44}$ & 85.26$_{\pm 0.97}$ & 81.02$_{\pm 0.28}$ & 85.93$_{\pm 0.06}$ \\
 & $\mathcal{P}_2$ + Synonym & 91.79$_{\pm 0.56}$ & 75.85$_{\pm 2.85}$ & 82.68$_{\pm 2.29}$ & 79.27$_{\pm 0.71}$ & 84.80$_{\pm 0.45}$ \\
 & \our + Synonym & \textbf{92.66$_{\pm 0.69}$} & \textbf{80.70$_{\pm 0.21}$} & \textbf{86.12$_{\pm 0.12}$} & \textbf{81.21$_{\pm 2.29}$} & \textbf{86.21$_{\pm 1.38}$} \\
\hline
\end{tabular}
\caption{Construction-strategy ablation of retriever (Recall@64, PC-POS) and reranker (Recall@1, MRR) results. We compare pseudo-pair construction strategies $\mathcal{P}_1$, $\mathcal{P}_2$, and their union \our, each combined with curated synonyms. \textbf{Bold} indicates the best result per column within each dataset.}
\label{tab:construction_ablation}
\end{table*}

\begin{table*}[h]
\centering
\small
\begin{tabular}{p{0.10\textwidth} p{0.62\textwidth} p{0.20\textwidth}}
\toprule
Ontology & Definition & Generated name \\
\midrule
NCBI & Autosomal dominant HEREDITARY CANCER SYNDROME in which a mutation most often in either BRCA1 or BRCA2 is associated with a significantly increased risk for breast and ovarian cancers. & Hereditary Breast and Ovarian Cancer Syndrome \\
\cmidrule(lr){2-3}
     & The presence in a cell of two paired chromosomes from the same parent, with no chromosome of that pair from the other parent. This chromosome composition stems from non-disjunction (NONDISJUNCTION, GENETIC) events during MEIOSIS. The disomy may be composed of both homologous chromosomes from one parent (heterodisomy) or a duplicate of one chromosome (isodisomy). & Uniparental Disomy \\
\cmidrule(lr){2-3}
     & Acquired, familial, and congenital disorders of SKELETAL MUSCLE and SMOOTH MUSCLE. & Muscular Diseases \\
\midrule
BC5CDR & A benzamide derivative that is used as a dopamine antagonist. & Tiapride Hydrochloride \\
\cmidrule(lr){2-3}
       & Pathologic processes that affect patients after a surgical procedure. They may or may not be related to the disease for which the surgery was done, and they may or may not be direct results of the surgery. & Postoperative Complications \\
\cmidrule(lr){2-3}
       & Used with drugs and chemicals for experimental human and animal studies of their ill effects. It includes studies to determine the margin of safety or the reactions accompanying administration at various dose levels. It is used also for exposure to environmental agents. Poisoning should be considered for life-threatening exposure to environmental agents. & toxicity \\
\midrule
CMO & Any measurement that deals with the process or development of a disease state, for example, the onset, progression or severity of the disease or its symptoms. & Disease process measurement \\
\cmidrule(lr){2-3}
    & The maximum arterial pressure within the cardiac cycle, i.e.\ at the point at which the heart is in its maximal state of contraction. This is the time when the blood is forced from the ventricles of the heart into the pulmonary artery and the aorta. & Systolic blood pressure \\
\cmidrule(lr){2-3}
    & Total distance around the body at the region of the body lateral to and including the hip joint or coxa. & Hip circumference \\
\midrule
VT & Any measurable or observable characteristic related to the physical magnitude of the portion of the body containing the brain and organs of sight, hearing, taste, and smell. & Head size trait \\
\cmidrule(lr){2-3}
   & The distance between the surfaces of the superficial layer of solid, hard bone that covers spongy bone. & Compact bone thickness \\
\cmidrule(lr){2-3}
   & The total proportion, quantity, or volume in milk of all inorganic elements or compounds that have importance in body functions. & Milk total mineral amount \\
\midrule
LPT & Any measurable or observable characteristic related to the proportion or amount in milk of the alpha fraction of the casein phosphoprotein. & Milk alpha-casein content \\
\cmidrule(lr){2-3}
    & Any measurable or observable characteristic related to the relative heaviness of the skeletal tissue of an animal following slaughter and removal of the head, digestive tract, internal organs, and possibly the hide and/or feet. & Dressed carcass bone weight \\
\cmidrule(lr){2-3}
    & Any measurable or observable characteristic related to the relative heaviness of the muscle tissue surrounding the femur of a bird. & Thigh muscle weight \\
\bottomrule
\end{tabular}
\caption{Three in-context examples used in the entity generation prompt for each ontology. Definitions are taken verbatim from the respective source ontology.}
\label{tab:gen-prompt-examples}
\end{table*}

\end{document}